\documentclass[journal]{IEEEtran}
\ifCLASSINFOpdf
\else
\fi

\usepackage{setspace}

\usepackage{graphicx}          
\usepackage[T1]{fontenc}
\usepackage[utf8]{inputenc}                     
\usepackage{array}
\usepackage{mathtools}
\usepackage{dcolumn}
\usepackage{amssymb}
\usepackage{amsmath}
\usepackage{booktabs}
\usepackage{bigstrut}

\usepackage{lmodern}      
\usepackage{algorithm}
\usepackage{algcompatible}
\usepackage{algpseudocode}
\usepackage{lineno}
\usepackage{color}
\usepackage{xspace}
\usepackage{amssymb}
\usepackage{float}
\usepackage{setspace} 
\usepackage{longtable}
\usepackage{graphics}
\usepackage{caption}
\usepackage{hyperref}
\usepackage{array}

\usepackage{lineno}
\usepackage{wasysym}
\usepackage{subcaption}
\usepackage{amsmath}
\usepackage{adjustbox}
\usepackage{multirow}
\usepackage{dblfloatfix}

\newcommand{\twoPBM}{\texttt{\textsc{Multi-stage ProBMoT}}} 
\newcommand{\PBMhh}{\texttt{\textsc{Single-stage ProBMoT}}}
\usepackage{ragged2e}
\usepackage[dvipsnames]{xcolor}

\begin{document}
%
\title{Equation Discovery \\ for Nonlinear System Identification}
%
%
%

\author{Nikola~Simidjievski,
        Ljup\v{c}o~Todorovski,
        Ju\v{s}~Kocijan,
        ~and~Sa\v{s}o~D\v{z}eroski

\thanks{All authors are affiliated with the Jo\v{z}ef Stefan Institute, Jamova cesta 39, 1000 Ljubljana, Slovenia.}
\thanks{N.~Simidjievski is also affiliated with the Department of Computer Science and Technology, University of Cambridge, Cambridge, United Kingdom. Email: nikola.simidjievski@ijs.si}
\thanks{L.~Todorovski is also affiliated with the Faculty of Administration, University of Ljubljana, Ljubljana, Slovenia. Email: ljupco.todorovski@fu.uni-lj.si}
\thanks{J.~Kocijan is also affiliated with the University of Nova Gorica, Nova Gorica, Slovenia. Email: jus.kocijan@ijs.si}
\thanks{S.~D\v{z}eroski is also affiliated with the Jo\v{z}ef Stefan International Postgraduate School, Ljubljana, Slovenia. Email: saso.dzeroski@ijs.si}
\thanks{\emph{Corresponding authors}: N.~Simidjievski and L.~Todorovski.}
\thanks{Last revision 28.03.2019.}
\thanks{Submitted to IEEE Transactions On Cybernetics}

}

%
%

\markboth{}%
{Simidjievski \MakeLowercase{\textit{et al.}}: Equation Discovery for Nonlinear System Identification}
%



\maketitle

\begin{abstract}
Equation discovery methods enable modelers to combine domain-specific knowledge and system identification to construct models most suitable for a selected modeling task. The method described and evaluated in this paper can be used as a nonlinear system identification method for gray-box modeling. It consists of two interlaced parts of modeling that are computer-aided. The first performs computer-aided identification of a model structure composed of elements selected from user-specified domain-specific modeling knowledge, while the second part performs parameter estimation.

In this paper, recent developments of the equation discovery method called process-based modeling, suited for nonlinear system identification, are elaborated and illustrated on two continuous-time case studies. The first case study illustrates the use of the process-based modeling on synthetic data while the second case-study evaluates on measured data for a standard system-identification benchmark. The experimental results clearly demonstrate the ability of process-based modeling to reconstruct both model structure and parameters from measured data.
\end{abstract}


%
\IEEEpeerreviewmaketitle

\section{Introduction}
%
%
%
%
\IEEEPARstart{D}{ata}-driven system identification \cite{Ljung1999} is omnipresent in many scientific and engineering domains: Given time-series data describing observed system behaviour, the task is to find a model, i.e., identify an appropriate model structure and estimate the values of the model parameters. Identifying the model structure is a challenging problem in many practical applications, especially in domains where structure inference from first principles is not an option. In such cases, one can opt for the black-box or the gray-box modeling paradigm. 
	
	In black-box modeling, we conjecture that the appropriate model structure belongs to a general class of structures, such as neural networks and fuzzy models \cite{Kerschen2006} or regression trees \cite{Aleksovski2015}. While nonlinear black-box models can achieve highly accurate reconstruction of the observed system behavior, they do not reveal the structure of the observed system. Namely, there is no clear correspondence between the assumed model structure (e.g., neural network) and the structure of the system at hand.
	
	Gray-box models, on the other hand, aim primarily at revealing the system structure. They are usually formalized as equations that scientists and engineers can comprehend and relate to \cite{Ljung2008}. To construct gray-box models, methods for symbolic regression \cite{Gray1998,SchmidtLipson2009} and equation discovery \cite{TodorovskiDzeroski2007} are commonly used. Symbolic regression addresses the gray-box modeling task with evolutionary methods, i.e., genetic algorithms \cite{SchmidtLipson2009}. Following these, equation structures are evolved from an initial set of variables, operators, functions and constants. Users have little or no control over the space of equation structures considered during search. Thus, the obtained models often include too complex and incomprehensible equations that make little contact with the modeling knowledge in the domain of use. Hence, the mapping between the structure of the model equations and the system structure is non-trivial and has to be inferred by human modelers.
	
	Equation discovery methods \cite{Langley1987}, on the other hand, allow the user to employ domain-specific knowledge and specify the appropriate space of candidate models. To this end, different formalisms for specifying equation fragments as components for building mathematical models have been proposed \cite{DzeroskiTodorovski2008}. While these formalisms can, in principle, be used in the context of evolutionary methods \cite{Mckay2010}, symbolic regression approaches rarely use them. Instead, equation discovery approaches often rely upon exhaustive or greedy search methods \cite{Todorovski2005HIPM}. The most recent equation discovery approach, process-based modeling (PBM), is based on a knowledge representation formalism that establishes an explicit mapping between the model structure and the structure of the observed system \cite{TodorovskiDzeroski2007,Bridewell2008}. It is closely related to component-based approaches for computer-aided design of electronic circuits \cite{Ashenden2008}, where models are composed from a library of standardized, well-characterized circuit components. These are with fixed constant parameters, used for manual model construction, where the modeler establishes a configuration of model components. PBM employs a library of similar components in the context of computer-aided input-output modeling, where the selection of an appropriate configuration of components (and the values of their parameters) is based on output-error fit against measured data.
	
	Process-based modeling has been applied in several practically relevant case studies. However, so far it has only been evaluated partially in limited setting and has not been systematically evaluated on a proper system identification benchmark yet. On one hand, PBM has been used to establish new models of aquatic ecosystems from data and domain-specific knowledge but with no reference to other known or well-established models \cite{Atanasova2006Bled}. On the other hand, PBM has been used to reconstruct the results of manual modeling efforts in the domain of systems biology \cite{Tanevski2016}. These efforts of establishing new and reconstructing existing models, used ProBMoT, a software tool for process-based modeling \cite{Cherepnalkoski2012}. Through the performed experiments, we have identified a number of design decisions that lead to an improved version of ProBMoT. 
	
	In this paper, we perform a systematic evaluation of the latest version of ProBMoT on a cascaded water-tank system \cite{WigrenSchoukens2013}, a well-known benchmark for nonlinear system identification. In particular, we analyze the robustness of PBM on synthetic data with different noise variances obtained by simulating the benchmark model. Furthermore, we also evaluate the PBM ability to reconstruct a valid model of the system dynamics from measurement data. Therefore, we present here a first proper evaluation of process-based modeling in the context of non-linear system identification.
	
	The remainder of the paper is organized as follows. Section~$\mathrm{II}$ discusses the relation of process-based modeling to alternative gray-box modeling approaches. Section~~$\mathrm{III}$ introduces the approach of process-based modeling by illustrating it on an example of modeling cascading water tanks. Section~$\mathrm{IV}$ elaborates the design of the modeling experiments, the results of which are reported and discussed in Section~$\mathrm{V}$. Finally, Section~$\mathrm{VI}$ concludes the paper with a summary and a brief outline of several directions for further research.

	\section{Related work}
	\subsection{Symbolic regression approaches}
	A number of gray-box modeling approaches to system identification aim at both structure and parameter identification of an observed system. Most notably, symbolic regression performed with genetic programming \cite{koza1994} has often been used to approach system identification \cite{Gray1998}. The basic idea of performing symbolic regression with genetic programming is to evolve populations of equations, where the fitness function being optimized measures the discrepancy between the observed data and the simulation of the equations. The equations are represented as tree structures, where the internal nodes correspond to mathematical operations and functions, while terminal nodes correspond to system variables and model parameters.
	
	The modeler can control the space of candidate models considered by genetic programming, by specifying the set of primitive operators and functions that can be used in the internal nodes of the evolving trees. In addition, the modeler can also specify a heuristic for model selection that goes beyond the usual degree-of-fit measure mentioned above. Madar et al.\cite{Madar2004} propose the use of parsimony principle when designing the fitness function by introducing a penalty for the complexity of the model equations. Rodriguez-Vazquez et al.\cite{Rodrigez2004} and Ferariu \& Patelli \cite{Ferariu2009} propose multi-objective approaches to follow the parsimony principle, where different aspects of degree-of-fit and model complexity are considered as objective functions.
	
	The common limitation of these methods is the coarse control given to the modeler over the space of candidate models. Often, modeling assumptions can not be easily or comprehensibly encoded as inputs to the genetic programming approaches: This leads to a risk of inferring models with implausible structure and/or parameter values. To address this issue, Whigham \& Recknagel \cite{WhighamRecknagel2001} and Cao et al. \cite{Cao2008} propose approaches where the space of model structures, considered by genetic programming, is constrained by using grammars \cite{Mckay2010}. While the grammars allow for flexible specification of the space of candidate models, their use is cumbersome and different for systems' modelers. In addition, the mapping between the structure of the obtained model and that of the observed system is non-trivial to establish: The inference of the mapping is left to the human modeler.
	
	\subsection{Component-based approaches}
	On the other hand, the mapping between the structure of the model and observed system is explicit in component-based approaches \cite{Ashenden2008}, where models are composed from a library of standardized, well-characterized circuit components. Note that the components have fixed constant parameter values, that correspond to physical circuit components and are used for manual construction and design of electrical circuits. No attempts have been made to use libraries of components in the context of computer-aided modeling from knowledge and input/output observational data. 
	
	While component-based approaches rely on physical components, block-oriented nonlinear approaches \cite{GiriBai2010} build models out of domain-independent linear, dynamic and non-linear, static components. In turn, such blocks can be combined into different structures, where each block models a certain aspect of the system dynamics. While some of these combinations are well known and correspond to physical phenomena \cite{Schoukens2017}, in principle, different combinations and variations of such components can yield a large number of different structures without a direct mapping to the structure of the observed dynamics at hand. Also, block-oriented approaches usually focus on the manual construction of models, where the block configuration is a priori prescribed by the modeler \cite{SchoukensJ2015}. M.Schoukens et al. \cite{Schoukens2014} and J.Schoukens et al. \cite{Schoukens2015} propose an approach that requires some modeler's decision regarding model construction by decomposing the structure identification task. Here, an initial model structure is established first by evaluating the performance of different combinations of linear components using best linear approximation against the nonlinear measured output. The identified initial structure, is then used by the modeler as the basis for subsequent steps of adding additional nonlinear blocks to complete model structure. 
	
	\subsection{Black-box approaches for gray-box modeling}
	Recently, some approaches to nonlinear black-box modeling have been also applied in the context of inferring gray-box models. In particular, Brunton et al. \cite{Brunton2016} combine linear regression with the construction of nonlinear terms and Lasso regularization to obtain nonlinear equations-based models that can be considered gray-box. The system SINDy introduces nonlinear terms by using a set of user-specified parameter-free transformations of the original system variables. These nonlinear terms are linearly combined with parameters estimated using simple linear regression. In this respect, SINDy is similar to the early equation discovery system LAGRANGE \cite{DzeroskiTodorovski1995}. A novelty in SINDy with the respect to LAGRANGE is the Lasso regularization, used to select terms that are relevant for modeling the observed data. Note however, that both methods are limited to inferring models with structure that is linear with respect to the parameters. Genetic programming, grammar-based equation discovery and process-based modeling are more general approaches that can infer models with structure that is nonlinear with respect to the model parameters.

	\section{Process-based modeling}
	
	Process-based modeling (PBM) is an approach to Equation Discovery that allows the modeler to formalize and use modeling knowledge specific to the domain of interest. We are going to introduce the PBM approach in the two following subsections. The first subsection introduces a general framework for representing modeling knowledge. In the second subsection, we present the method that employs the formalized knowledge to discover continuous-time models from observation data.
	
	\subsection{Formalization of the modeling knowledge}
	
	The PBM framework for formal representation of modeling knowledge is based on two general concepts of entities and processes. Entities correspond to the static components of the observed system, while processes relate to the dynamic interactions among entities. Each entity includes variables and constants that denote time-varying and time-invariant properties of the corresponding system component, respectively. While entity constants correspond to constant (time-invariant) model parameters, entity variables correspond to the state, input and output variables of the model. The system state is therefore represented with the set of the state variables included in the model entities.	On the other hand, each process specifies the interacting entities, process constants and equations. The latter provide the mathematical model of the process influence on the interacting entities, more specifically, on the temporal change, i.e., time derivatives of their state variables. Each process equation can include constants and both system and input/output variables of the interacting entities as well as process constants. When several processes influence the same system variable, their mathematical models are summed up to obtain the equation for modeling the dynamics of the particular variable.
	
	\begin{figure*}
		\includegraphics[width=\textwidth]{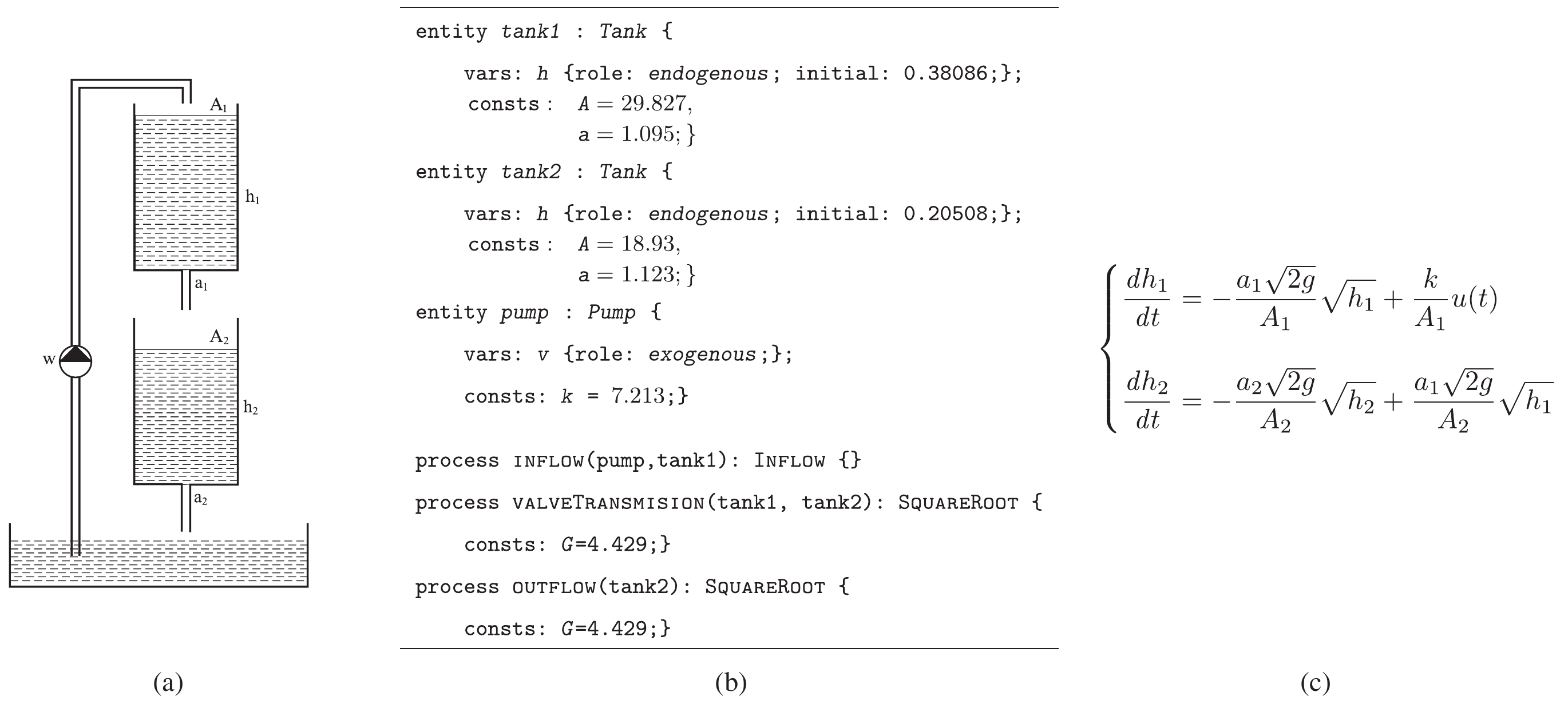}
		\caption{(a) The diagram of the three components of the water-tanks system; (b) The process-based model of the water-tanks system; (c) The equation-based model of the water-tanks system.}
		\label{fig:pbm-wt}%
	\end{figure*}
	
	Figure~\ref{fig:pbm-wt} provides an illustrative example of a process-based model of a dynamic system consisting of two cascaded water tanks. It consists of three entities and three processes as shown in Figure~\ref{fig:pbm-wt}b. The three entities correspond to the three system components depicted in Figure~\ref{fig:pbm-wt}a: the two tanks and the water pump that fills the upper tank. Each {\tt tank} entity includes a single state (endogenous) variable {\tt h}, corresponding to the height of the water level in the tank, as well as two constants {\tt A} and {\tt a} denoting the areas of the tanks and their effluent areas, respectively. The third entity {\tt pump} includes the input (exogenous) variable {\tt v} representing the input voltage signal $ u(t) $ applied to the pump: the higher the voltage -- the higher the water inflow in the upper tank. The voltage-to-flow conversion constant of the pump is denoted with {\tt k}.
	The three processes correspond to the entity interactions, which, in this particular example, represent the water flow between the tanks and the pump. The first process {\tt\sc inflow} models the flow from the pump to the upper tank, the second process {\tt\sc valveTransmission} models the flow from the upper to the lower tank, and the third models the {\tt\sc outflow} of water from the lower tank and the system. The latter two processes assume a square-root influence of the height of the water in the tanks on the intensity of the tank valve outflow. The single constant parameter of these two processes specifies the valve transition constant {\tt G} that equals $ \sqrt{2 g} $ (where $g$ is gravitational acceleration).
	
	Process-based models provide additional structure to the model equations with a modular decomposition of the equations into building blocks. These correspond to the knowledge about building models in the domain of interest. In the particular example, the correspondence between the physical structure of the system and the process-based model structure is obvious and explicit. The benefits of this property are twofold. On one hand, it improves the descriptive power of the equations by adding an explanatory layer on top of them. In particular, the equations in Figure~\ref{fig:pbm-wt}c do not explicitly reveal the meaning of the repetitive expression $ a_1 \sqrt{2 g} \sqrt{h_1} $. In contrast, the process-based model ties that expression to the flow from the upper to the lower tank. On the other hand, and even more importantly, the decomposition into building blocks provides means for formulating the task of inducing process-based model from data as a combinatorial optimization task. The latter is then constrained to those models that have plausible structure and are therefore acceptable to domain experts. Note, finally, that the process-based models can be easily and automatically rewritten into the standard mathematical form of differential equations that allows for model simulation and analysis.
	
	The process-based formalism for representing modeling knowledge follows the model decomposition into building blocks. In particular, a modeling expert can specify the set of available building blocks in the domain of interest as a library of entity and process templates. The templates represent generic building blocks that, in a particular modeling scenario, can be instantiated into specific components of the model of the observed system. Each entity template prescribes the set of constant and variable properties important for modeling the corresponding static system component in the domain. Similarly, the process templates are generic interactions and prescribe template mathematical expressions for modeling them.
	
	\begin{table}[b!]
		\caption{Library of template entities and processes for modeling an arbitrary configuration of water tanks with three alternative valve transmission functions and three alternative outflow functions.}
		\centering
		\ttfamily\footnotesize 
		\begin{tabular}{l}
			\hline
			\color{red}template entity \textsl{Tank} \{ \\
			\color{red}\qquad	vars: \textsl{h} \{aggregation:sum, range:<0,500>\};\\
			\color{red}$\begin{aligned}
			\qquad \texttt{consts} : \quad  &\textsl{A} \quad \texttt{\{range:<1.0E-3,30>\}}, \\
			&\textsl{c}\quad \texttt{\{range:<1.0E-3,30>\}};\}\\
			\end{aligned}$\\
			\color{BlueViolet}template entity \textsl{Pump} \{ \\
			\color{BlueViolet}\qquad	vars: \textsl{v};\\
			\color{BlueViolet}\qquad	consts: \textsl{k} \{range:<1.0E-3,30>\};\}\\
			\\
			\color{orange} template process \textsc{Inflow}(\textsl{p}: \textsl{Pump}, \color{orange}\textsl{t}: \textsl{Tank})\{ \\
			\color{orange}\qquad 	equations: $\mathrm{td(t.h) =  p.k \cdot p.v / t.A}$;
			\color{orange}\} \\ 
			
			\color{RoyalBlue}template process \textsc{ValveTransmission}(\textsl{t1}: \textsl{Tank}, \textsl{t2}: \textsl{Tank}) \{ \\
			
			\color{RoyalBlue}\qquad consts : \textsl{G} \{range:<0,10>\}; //$sqrt(2g)$				
			\color{RoyalBlue}\}\\
			\color{RoyalBlue}template process \textsc{\color{Blue}{\textbf{S}}\color{RoyalBlue}quareRoot}:\texttt{\textsc{ValveTransmission}} \{ \\
			\color{RoyalBlue}$\begin{aligned}
			\qquad \texttt{equations} : \quad & \mathrm{td(t1.h) = - G \cdot pow(t1.h, 0.5)\cdot t1.a/t1.A},\\ 
			& \mathrm{td(t2.h) = G \cdot pow(t1.h, 0.5)\cdot t1.a/t2.A};\}\\
			\end{aligned}$\\
			\color{RoyalBlue}template process \textsc{\color{Blue}{\textbf{L}}\color{RoyalBlue}inear}:\texttt{\textsc{ValveTransmission}} \{ \\
			\color{RoyalBlue}$\begin{aligned}
			\qquad \texttt{equations} : \quad & \mathrm{td(t1.h) = - G \cdot t1.h \cdot t1.a/t1.A},\\ 
			& \mathrm{td(t2.h) = G \cdot t1.h \cdot t1.a/t2.A};\}\\
			\end{aligned}$\\
			\color{RoyalBlue}template process \textsc{\color{Blue}{\textbf{E}}\color{RoyalBlue}xponential}:\texttt{\textsc{ValveTransmission}} \{ \\
			\color{RoyalBlue}$\begin{aligned}
			\qquad \texttt{equations} : \quad & \mathrm{td(t1.h) = - G \cdot exp(t1.h)\cdot t1.a/t1.A},\\ 
			& \mathrm{td(t2.h) = G \cdot exp(t1.h)\cdot t1.a/t2.A};\}\\
			\end{aligned}$\\

			\color{Green}template process \textsc{Outflow}(\textsl{t}: \textsl{Tank}) \{ \\
			\color{Green}\qquad consts : \textsl{G} \{range:<0,10>\};\\	
			\color{Green}template process \textsc{\color{JungleGreen}{\textbf{S}}\color{Green}quareRoot}:\texttt{\textsc{Outflow}} \{ \\
			\color{Green}\qquad 	equations: $\mathrm{td(t.h) = - G \cdot pow(t.h, 0.5) \cdot t.a/t.A }$;\} \\ 
			\color{Green}template process \textsc{\color{JungleGreen}{\textbf{L}}\color{Green}inear}:\texttt{\textsc{Outflow}} \{ \\
			\color{Green}\qquad 	equations: $\mathrm{td(t.h) = - G \cdot t.h  \cdot t.a/t.A}$;\} \\ 
			\color{Green}template process \textsc{\color{JungleGreen}{\textbf{E}}\color{Green}xponential}:\texttt{\textsc{Outflow}} \{ \\
			\color{Green}\qquad 	equations: $\mathrm{td(t.h) = - G \cdot exp(t.h) \cdot t.a/t.A}$;\} \\ 
			\hline
			
		\end{tabular}%
		\label{tab:pbmlib-wt}%
	\end{table}%
	
	Table~\ref{tab:pbmlib-wt} provides an example library of template entities and processes for modeling systems of connected water tanks and pumps. The two generic entities of {\tt Tank} and {\tt Pump} can represent an arbitrary component of any system in the water-tanks domain. The entities {\tt tank1} and {\tt tank2} in the example process-based model from Figure~\ref{fig:pbm-wt}b represent two specific instances of the template entity {\tt Tank}. Note that the modeling knowledge encoded in the template entities include specifications of the constant and variable properties of the corresponding system components with their operational ranges rather than specific values. The {\tt aggregation:~sum} declaration specifies that when an entity {\tt Tank} participates in more than one process, the multiple influences of the different processes on the entity variable {\tt h} are summed up.
	
	The template processes are organized in a hierarchy. In Table~\ref{tab:pbmlib-wt}, the {\tt\sc ValveTransmission} template represents the root node of a hierarchy with three descendants, i.e. {\tt\sc SquareRoot}, {\tt\sc Linear} and {\tt\sc Exponential}, that correspond to three alternative mathematical models of valve transmission. The first assume the square-root, the second linear and the third exponential influence of the water level in the tank on the tank-valve outflow. Similarly, the {\tt\sc Outflow} hierarchy of template processes specifies three alternative models of water flow from the system to its environment.
	
	The process-based library provides a general recipe for building models in the domain of use. Following the encoded knowledge, models of arbitrary complexity can be established: They can include an arbitrary number of entities and interactions among them. In particular, the library from Table~\ref{tab:pbmlib-wt} can be re-used for modeling water-tanks systems with an arbitrary number and configurations of tanks, pumps and flows among them and the system environment. Such knowledge also allows the modeler to explore various modeling assumptions. In situations where the valve transmission function is known, one can prescribe that the specific process should be an instance of, e.g., the {\tt\sc SquareRoot} template. On the other hand, if the valve transmission function is unknown and should be induced from the observational data, one should prescribe the root template of the hierarchy, i.e., {\tt\sc ValveTransmission}.
	
	The alternative configurations of model components and corresponding mathematical expressions can be automatically explored by the algorithm for inducing process-based models from data and knowledge, introduced in the following subsection.
	
	\subsection{Inducing process-based models}
	
\begin{figure*}
		\centering
		\includegraphics[width=0.8\textwidth]{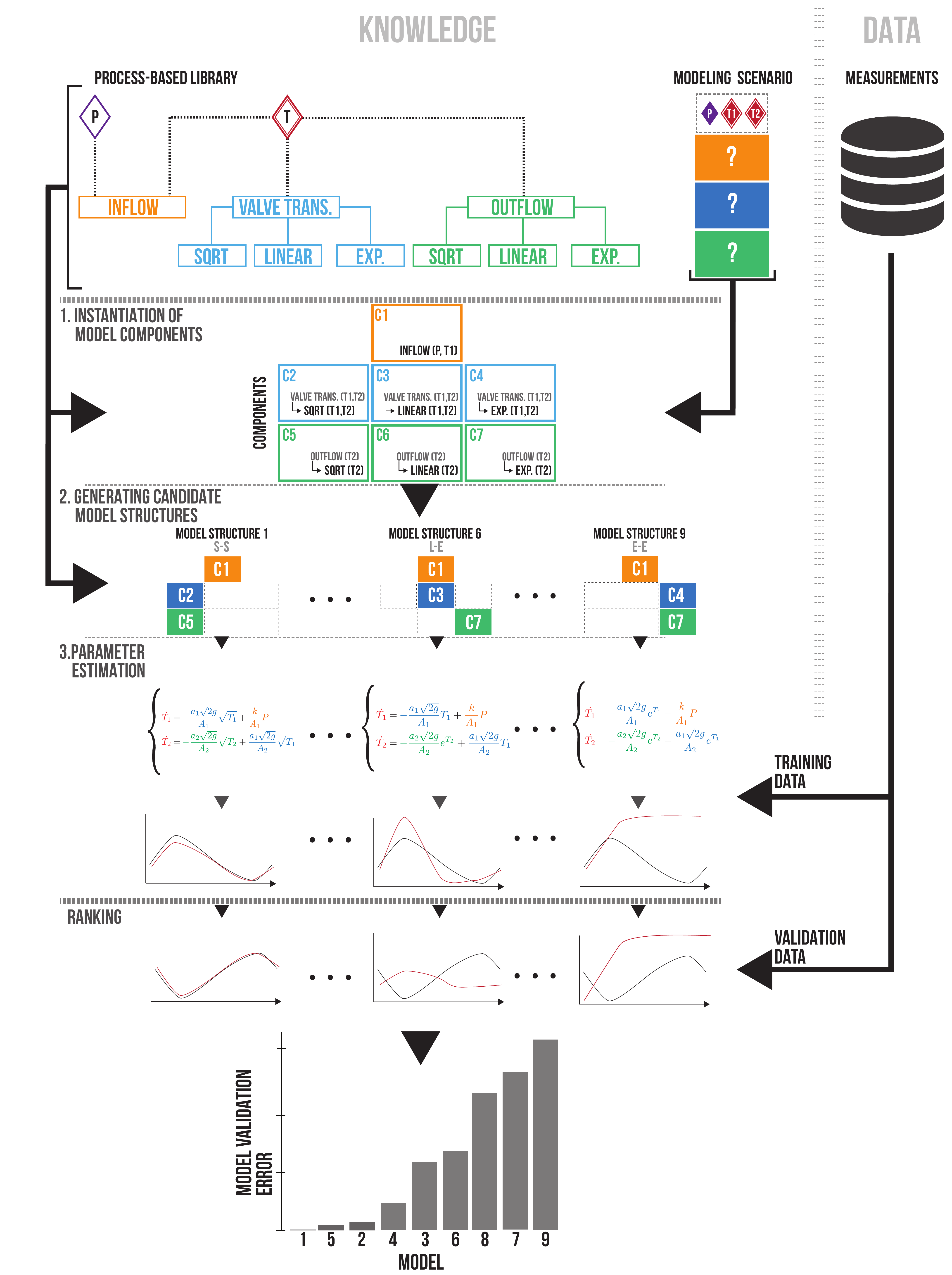}
		\caption{Workflow of the ProBMoT algorithm for inducing process-based models from knowledge and data. Given the process-based library and the modeling scenario, 9 candidate models structures are constructed, listed in Table~\ref{tab:Legend}. These structures are mapped to equations, have their parameters estimated and are finally ranked based on model validation error.}
		\label{fig:probmot}%
	\end{figure*}

	Figure~\ref{fig:probmot} outlines the ProBMoT algorithm for inducing process-based models, which takes three inputs. The first input is the process-based library introduced in the previous subsection. The second input specifies the modeling scenario including the specific entities that appear in the observed system, the known configuration of the interactions among them and the modeling assumptions. The third is the training and validation data sets of measurements of the state and input/output variables of the observed system. Given its inputs, ProBMoT proceeds in three steps: (1) instantiating the templates from the library into specific model components, (2) generating the candidate model structures corresponding to combinations of the model components, and (3) estimating the constant parameters and the fit of each model structure against data. Finally, the output of ProBMoT is a list of continuous-time process-based models ranked according to their performance, e.g. the discrepancy between the model simulation and the validation data set.
	
	In the first step, given the entities of the observed system and the modeling assumptions, each template process is instantiated into a number of specific instance processes. For example, given two tank entities, the template process {\tt\sc ValveTransmission} has six instances. The first three of them correspond to the situation where the water flows form the first to the second tank, with each of the three corresponding to one valve transmission form. The remaining three correspond to the flows from the second task to the first. Table~\ref{tab:Incomplete} introduces a specific modeling scenario for an observed system consisting of a cascade of a pump and two water tanks: the pump fills the first tanks, there is a flow between the first and the second tank, and the water of the second tank outflows into the system environment. In this particular scenario, the {\tt\sc Inflow} generic process has a single instance and each of the {\tt\sc ValveTransmission} and {\tt\sc Outflow} process has three instances.
	
	\begin{table}[]
		\caption{Modeling scenario for the system of two cascaded water tanks and a pump.}
		\centering
		\ttfamily 
		\begin{tabular}{>{\footnotesize  }l}
			\hline
			\color{red}entity \textsl{tank1} : \textsl{Tank} \{ \\
			\color{red}\qquad vars: \textsl{h} \{role: \textsl{endogenous}; initial: 0.38086;\};\\
			\color{red}$\begin{aligned}
			\qquad \texttt{consts} : \quad  &\textsl{A}= \texttt{null}, \\
			&\textsl{a} = \texttt{null}; \}\\
			\end{aligned}$\\
			\color{red}entity \textsl{tank2} : \textsl{Tank} \{ \\
			\color{red}\qquad vars: \textsl{h} \{role: \textsl{endogenous}; initial: 0.20508;\};\\
			\color{red}$\begin{aligned}
			\qquad \texttt{consts} : \quad  &\textsl{A}= \texttt{null}, \\
			&\textsl{a} = \texttt{null}; \}\\
			\end{aligned}$\\
			\color{BlueViolet}entity \textsl{pump} : \textsl{Pump} \{ \\
			\color{BlueViolet}\qquad vars: \textsl{v} \{role: \textsl{exogenous};\};\\
			\color{BlueViolet}\qquad consts: \textsl{k} = null;\}\\
			\\
			\color{orange}process \textsc{inflow}(pump,tank1): \textsc{Inflow} \{\} \\
			\color{RoyalBlue}process \textsc{valveTransmision}(tank1, tank2):\\
			\color{RoyalBlue} \qquad\textsc{ValveTransmision} \{
			consts: \textsl{G}=4.429;\}  \\
			\color{Green}process \textsc{outflow}(tank2): \textsc{Outflow} \{
			consts: \textsl{G}=4.429;\}\\
			
			\hline
		\end{tabular}%
		\label{tab:Incomplete}%
	\end{table}%
	
	The seven process instances represent components for building process-based models of the observed system in the second step of the algorithm. Finding an optimal model is equivalent to the task of finding an optimal combination of the model components. This is a task of combinatorial optimization, which can be approached by an arbitrary algorithm for combinatorial search. In the particular modeling scenario, from the seven model components, we need to compose a model that has an {\tt\sc Inflow}, {\tt\sc ValveTransmission} and an {\tt\sc Ouflow} component. This leads to $ 1 \times 3 \times 3 = 9 $ candidate combinations, so we employ exhaustive search, i.e., enumerate all the candidates. Alternatively, when facing more complex modeling scenarios leading to large search spaces, one can employ scalable method for combinatorial optimization that perform incomplete, heuristic search through the space of candidate combinations. To this end, symbolic regression methods often employ evolutionary approaches, such as genetic algorithms \cite{SchmidtLipson2009}, while equation discovery methods employ greedy search algorithms \cite{Todorovski2005HIPM}.

	\begin{table*}[h]
		\centering
		\caption{The list of nine candidate model structures considered by ProBMoT. The acronyms correspond to the different components included in the respective interaction. For e.g,  Model S-S denotes a model structure with {\tt\sc SquareRoot} modeling alternative for both processes {\tt\sc ValveTransmission} and {\tt\sc Outflow}. All model structure include the {\tt\sc Inflow} process.}
		\scalebox{.69}{
		\begin{tabular}{l|rrrrrrrrr}
			\hline
			\hline
			{\tt\sc ValveTransmision}  &  {\tt\sc SquareRoot}  &  {\tt\sc SquareRoot}  &  {\tt\sc SquareRoot}  &  {\tt\sc Linear} & {\tt\sc Linear} & {\tt\sc Linear} & {\tt\sc Exponential}   & {\tt\sc Exponential}   & {\tt\sc Exponential} \bigstrut[t]\\
			{\tt\sc Outflow} &  {\tt\sc SquareRoot}  & {\tt\sc Linear} & {\tt\sc Exponential}   &  {\tt\sc SquareRoot}  & {\tt\sc Linear} & {\tt\sc Exponential}   &  {\tt\sc SquareRoot}  & {\tt\sc Linear} & {\tt\sc Exponential} \bigstrut[b]\\
			\hline
			Model & S-S    & S-L    & S-E    & L-S    & L-L    & L-E    & E-S    & E-L    & E-E \bigstrut\\
			\hline
			\hline
		\end{tabular} }
		\label{tab:Legend}%
	\end{table*}%
	In the final, third step of the algorithm, ProBMoT evaluates each candidate model structure by estimating the values of its constant parameters. Recall that the list of constant parameters and the ranges of their values are included in the process-based library. ProBMoT approaches the parameter estimation as a numerical optimization task. The objective function for numerical optimization is the output error, i.e., the discrepancy between the simulated model response and the measured system response in the training data set \cite{Breiman1984}, defined as:
	
	\begin{equation}
	\label{eq:rermse}
	{\it RRMSE}(y) = \sqrt{\frac{\sum_{t=0}^{n}(y_{t}-\hat{y_{t}})^{2}}{\sum_{t=0}^{n}(\bar{y}-\hat{y_{t}})^{2}}}
	\end{equation}
	
	\noindent
	
	In Eq.~(\ref{eq:rermse}) $ n $ denotes the number of measurements, $y_{t}$ and $\hat{y_{t}}$ correspond to the {\it measured} and {\it simulated} values of the output variables $ y $ and $ \bar{y} $ denotes the mean value of $ y $ in the data set. ProBMoT employs the CVode library~\cite{CohenHindmarsh1996} for simulating continuous-time models in the form of ordinary differential equations. CVode implements a general-purpose ODE solver with linear multi-step variable-coefficient methods for integration. Note that we perform long-term simulation of the model equations given the initial conditions only: This is in contrast to most simulation approaches that rely on value-corrections of the simulation response at each simulation step~\cite{Zhang2004}. Given that the objective function includes simulation of non-linear continuous-time models and an output-error objective function, ProBMoT employs nonlinear methods for numerical optimization. In particular, it uses the Differential Evolution algorithm as implemented in the meta-heuristic optimization framework jMetal~\cite{Durillo2011}. Differential evolution is arguably a powerful stochastic optimization algorithm with encouraging results related to its asymptotic convergence towards a global optima~\cite{Ghosh2012}.
	
	\section{Experimental design}
	
	
	To properly test the hypothesis that the process-based modeling approach can accurately perform nonlinear system identification, we propose investigating the two aspects of: (1) structure identification and (2) parameter estimation. To this end, the experiments consider modeling a cascading water tank system with two tanks as shown in Figure~\ref{fig:pbm-wt}a, where we focus on modeling the dynamics of the lower tank $h_{2}$. In the first line of experiments, we investigate ProBMoT's ability to discriminate among different instantiated candidate structures. Next, we analyze its ability to correctly reconstruct the values of the model parameters. 
	
	In particular, we consider two modes of operation of ProBMoT: (1) multi-stage mode -- \twoPBM{} and (2) single-stage mode -- \PBMhh{}. The former, relates to a standard approach in system identification where the problem is decomposed so that the dynamics of each signal is modeled separately \cite{Schoukens2015}. Namely, using ProBMoT we first attempt to establish the structure and parameters of model of the first tank, using the signal $h_{1}$ in the learning process. In turn, the best performing (intermediate) model is used for establishing the structure and parameters of the lower tank using the signal $h_{2}$. On the other hand, \PBMhh{}, refers to the standard operation of ProBMoT, where it attempts to learn the whole model for both signals $h_{1}$ and $h_{2}$ simultaneously. 
	
	Regarding the data, in our experiments we consider both synthetic data, by varying the values of the signals $h_{1}$ and $h_{2}$ at each time point, as well as measured data \cite{WigrenSchoukens2013}. The latter consists of 2500 measurements of the input signal $u(t)$ and the two output signals, i.e., the water levels of the two tanks, $h_{1}$ and $h_{2}$. To properly evaluate model performance, we split these data into a train, a validation, and a test set, consisting of 1000, 500 and 1000 measurements, respectively. In particular, we learn the model structure and parameters using the train data, we select an optimal model with respect to its performance on a validation data set, and evaluate the performance of the selected model on the test set. 
	
	The synthetic data is generated by simulating the model from Fig~\ref{fig:pbm-wt}b using arbitrarily selected parameter values of $a_{1}=0.65$, $A_{1}=20$, $a_{2}=0.7$, $A_{2}=12$ and $k=5$. For the input signal $u(t)$, we used the real measurements, while for $h_1$ and $h_2$, we used the simulated responses with Gaussian noise added with the equation $ y_{noise}(t) = y(t) \cdot (1 + N(0,v)) $, where $ N(0,v) $ denotes a normal-distribution random variable with mean 0 and variance $ v $. Note that we do not add noise to the input signal, since it comes from the measurements. We generate five different sets of synthetic responses with variance of 0.01, 0.02, 0.05, 0.1 and 0.2, where the uncorrupted responses correspond to variance 0. This yields six different synthetic experimental cases/data sets.
	
	In both the single-stage and the multi-stage modes, ProBMoT uses the process-based library of domain knowledge for modeling cascading-tank dynamics, presented in Table~\ref{tab:pbmlib-wt}. The library combined with the modeling scenario given in Table~\ref{tab:Incomplete}, yields 9 candidate model structures shown in Table~\ref{tab:Legend}. For instance, model S-S denotes a model where the processes {\tt\sc ValveTransmission} and {\tt\sc Outflow} have sub-linear {\tt\sc SquareRoot} dynamics. Note that, the artificial data was generated with the model S-S with the parameter values stated in the previous paragraph. The different modeling scenarios used in the experiments with single-stage and the multi-stage ProBMoT are given in Appendix~A.
	
	Furthermore, the following parameter settings for Differential Evolution were used: a population size of 60, strategy {\sl rand/1/bin}, differential weight ($ F $) and a crossover probability ($ Cr $) of 0.9. The limit on the number of evaluations of the objective function is $5\cdot10^4$ per parameter. The particular parameter setting of DE is based on previous studies~\cite{Tashkova2012}. Finally, for simulating the models, we used the CVode simulator~\cite{CohenHindmarsh1996} with a standard setting of absolute and relative tolerances, ${10}^{-8}$ and ${10}^{-4}$, respectively. For generating the synthetic data as well as perform model simulation in ProBMoT, we use the backward differentiation method combined with Newton iteration and a preconditioned Krylov method with ${10}^{5}$ steps implemented in \cite{CohenHindmarsh1996}.
	
	\section {Results}
	
	We investigate the ability of ProBMoT to reconstruct the correct models in both a synthetic and a real-world setting. In the first line of experiments, we examine ProBMoT's ability to recover the ground-truth model S-S from synthetic data. In particular, we present performance comparisons between the two modes of ProBMoT, \twoPBM{} and \PBMhh{}, in terms of $RRMSE_{validation}$ and $RRMSE_{test}$ on the six synthetic experimental cases. Next, we investigate the ability of the two modes of ProBMoT to estimate the ground-truth parameters from the synthetic data. In the second line of experiments, we test the ability of ProBMoT to identify the structure and parameter of the model from measurements of the water-tanks system. 
	
	\subsection{Results on synthetic data}
	
\begin{figure}[b!]
		
		\centering
		\begin{subfigure}[t]{0.49\linewidth}
			\includegraphics[width=\linewidth]{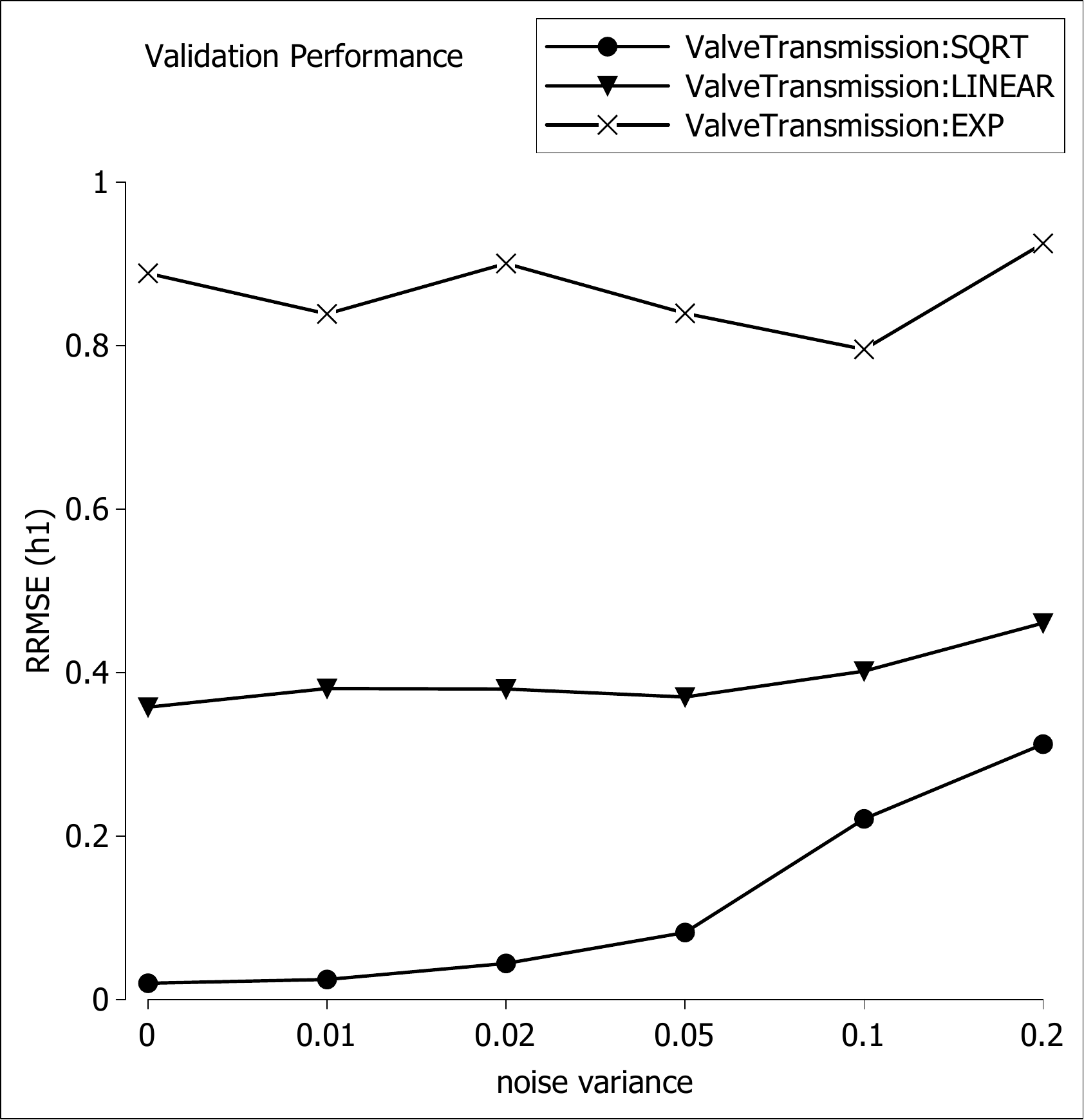}
			\subcaption{}
		\end{subfigure}
		\begin{subfigure}[t]{0.49\linewidth}
			\includegraphics[width=\linewidth]{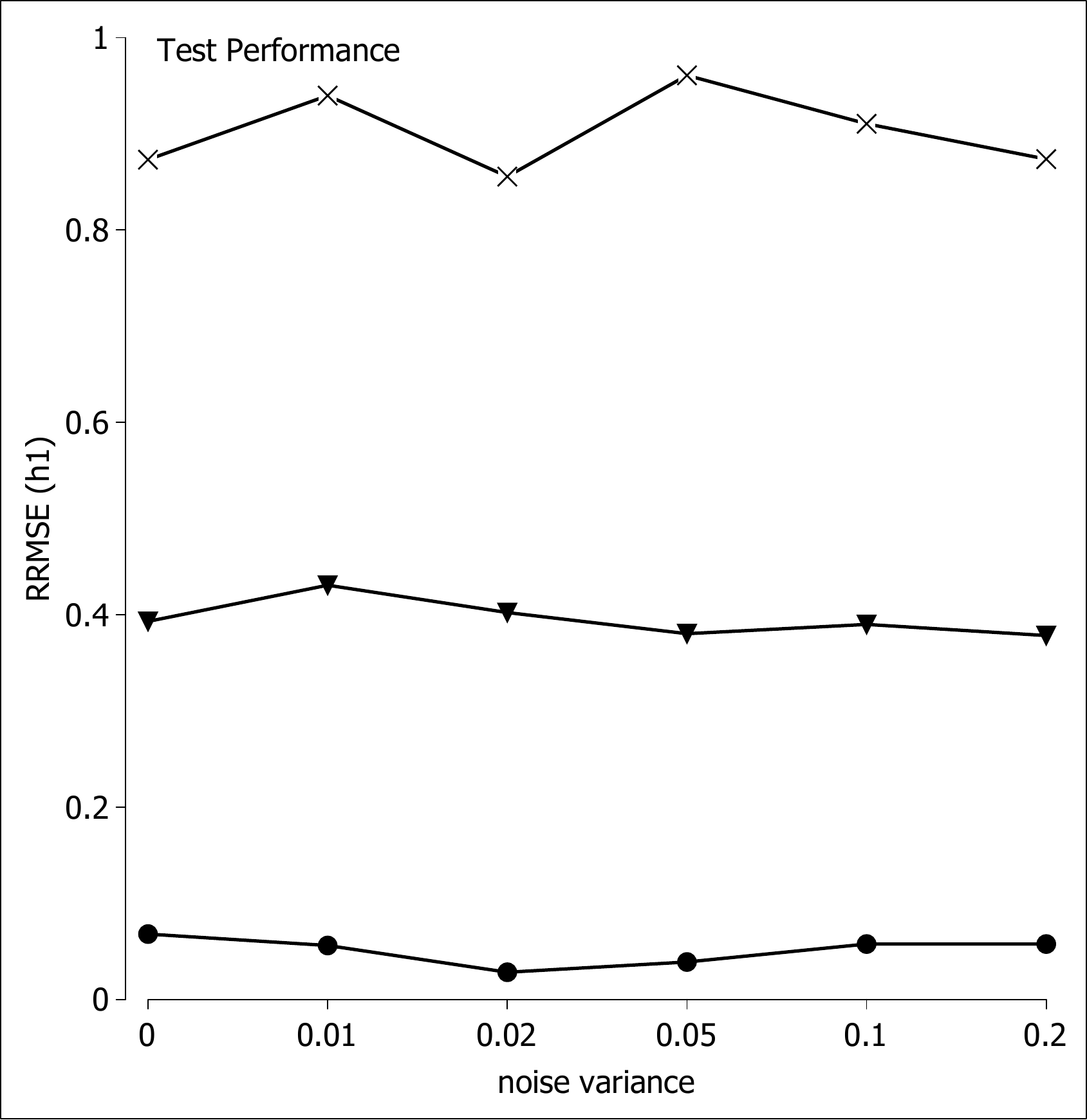}
			\subcaption{}
		\end{subfigure}
		
		\begin{subfigure}[t]{0.49\linewidth}
			\includegraphics[width=\linewidth]{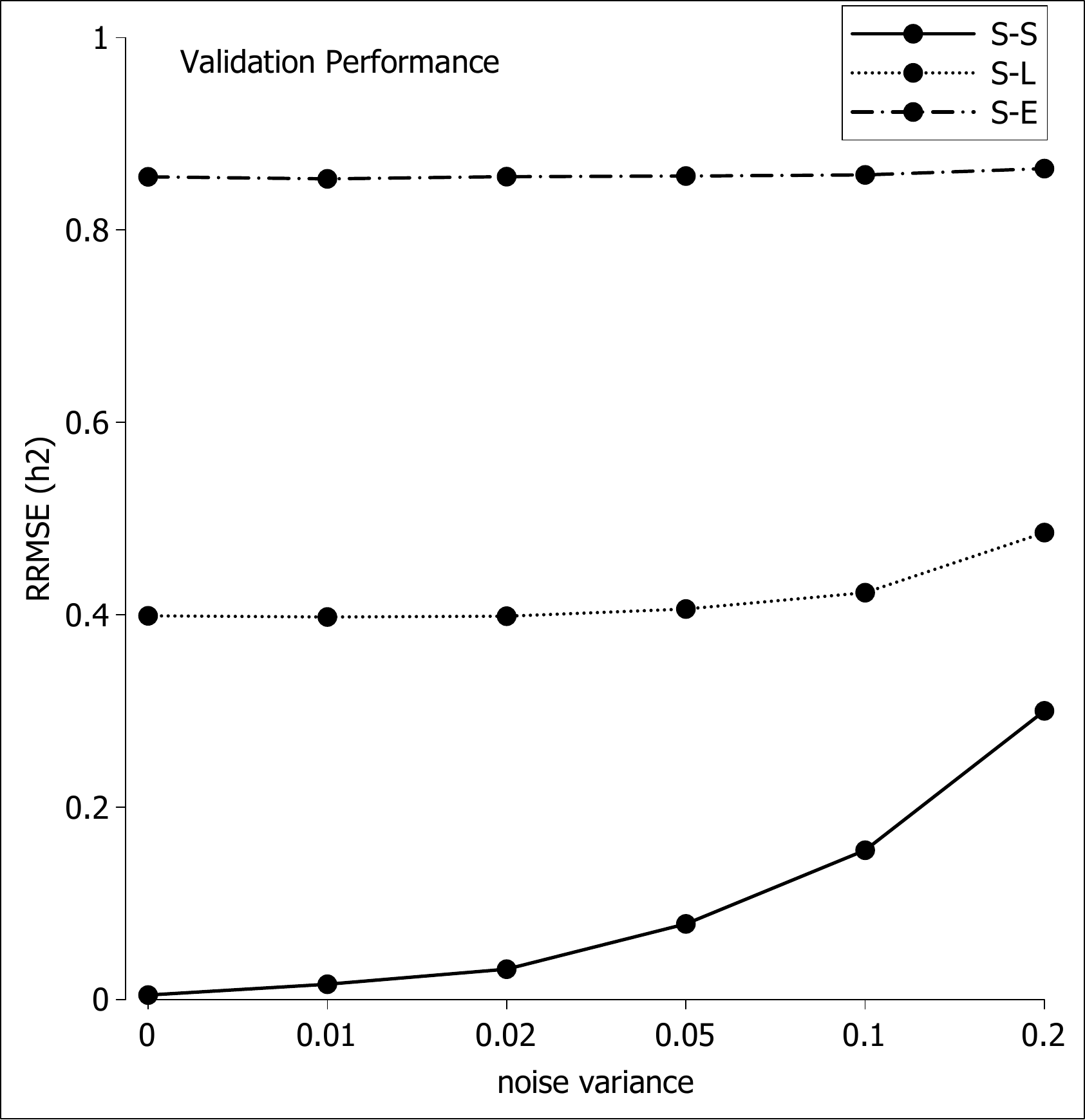}
			\subcaption{}
		\end{subfigure}
		\begin{subfigure}[t]{0.49\linewidth}
			\includegraphics[width=\linewidth]{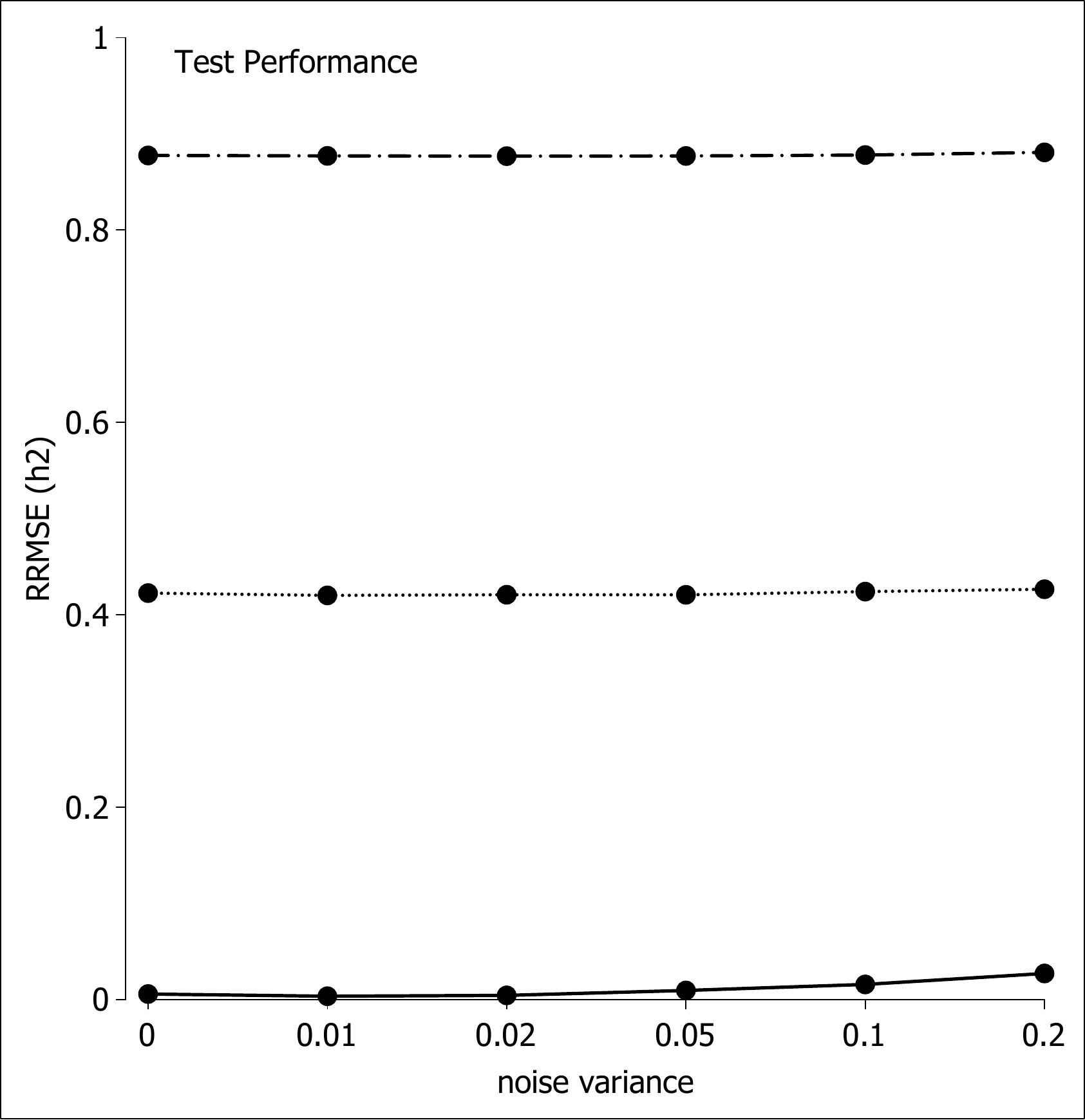}
			\subcaption{}
		\end{subfigure}

		\caption{\twoPBM{}: \textbf{Top}- Performance comparison of the simulation error ($RRMSE$) of the  $h_{1}$ signal obtained from the 3 intermediate model structures on the (a) validation and (b) test synthetic data sets in the six different synthetic cases. \textbf{Bottom}- Performance comparison of the simulation error ($RRMSE$) of the output signal $h_{2}$ obtained from the remaining 3 model structures on the (c) validation and (d) test synthetic data sets in the six different synthetic cases. \twoPBM{} can correctly identify both the intermediate model structure from the signal  $h_{1}$ in the first stage, as well as the complete model structure in the second stage.}
		\label{fig:2StageNoise}%
	\end{figure}

	Figure~\ref{fig:2StageNoise} depicts the errors of the models learned with \twoPBM{}. The top two graphs in Figure~\ref{fig:2StageNoise} refer to the errors of the three "intermediate" candidate models of the dynamics of $h_{1}$ considered in the first stage. Figure~\ref{fig:2StageNoise}a shows that ProBMoT can identify the structure of the process {\tt\sc ValveTransmission} for $h_{1}$ correctly. Moreover, Figure~\ref{fig:2StageNoise}b shows that the identified intermediate model with a {\tt\sc SquareRoot} alternative of the process {\tt\sc ValveTransmission} for the top tank has a constant and substantially better test performance than the other two models (with the {\tt\sc Linear} or {\tt\sc Exponential} alternative).

	The two bottom graphs in Figure~\ref{fig:2StageNoise} depict the errors of the three final models that complete the correctly identified intermediate model of $h_1$ in the first stage. Figure~\ref{fig:2StageNoise}c shows that ProBMoT selects the correct complete model S-S for all noise variances. In particular, the validation error of the model S-S in the noise-free scenario is almost $40$ percentage points lower than the second ranked model S-L, and almost $20$ percentage points lower for the 0.2 noise variance. This result is also in-line with the test performance of the S-S model, depicted in Figure~\ref{fig:2StageNoise}d, which, in general, performs up to $30$ percentage points better than the second ranked model S-L and up to $90$ percentage points better than the third-ranked S-E.
	
\begin{figure}[]
		\centering
		\begin{subfigure}[t]{0.49\linewidth}
			\includegraphics[width=\linewidth]{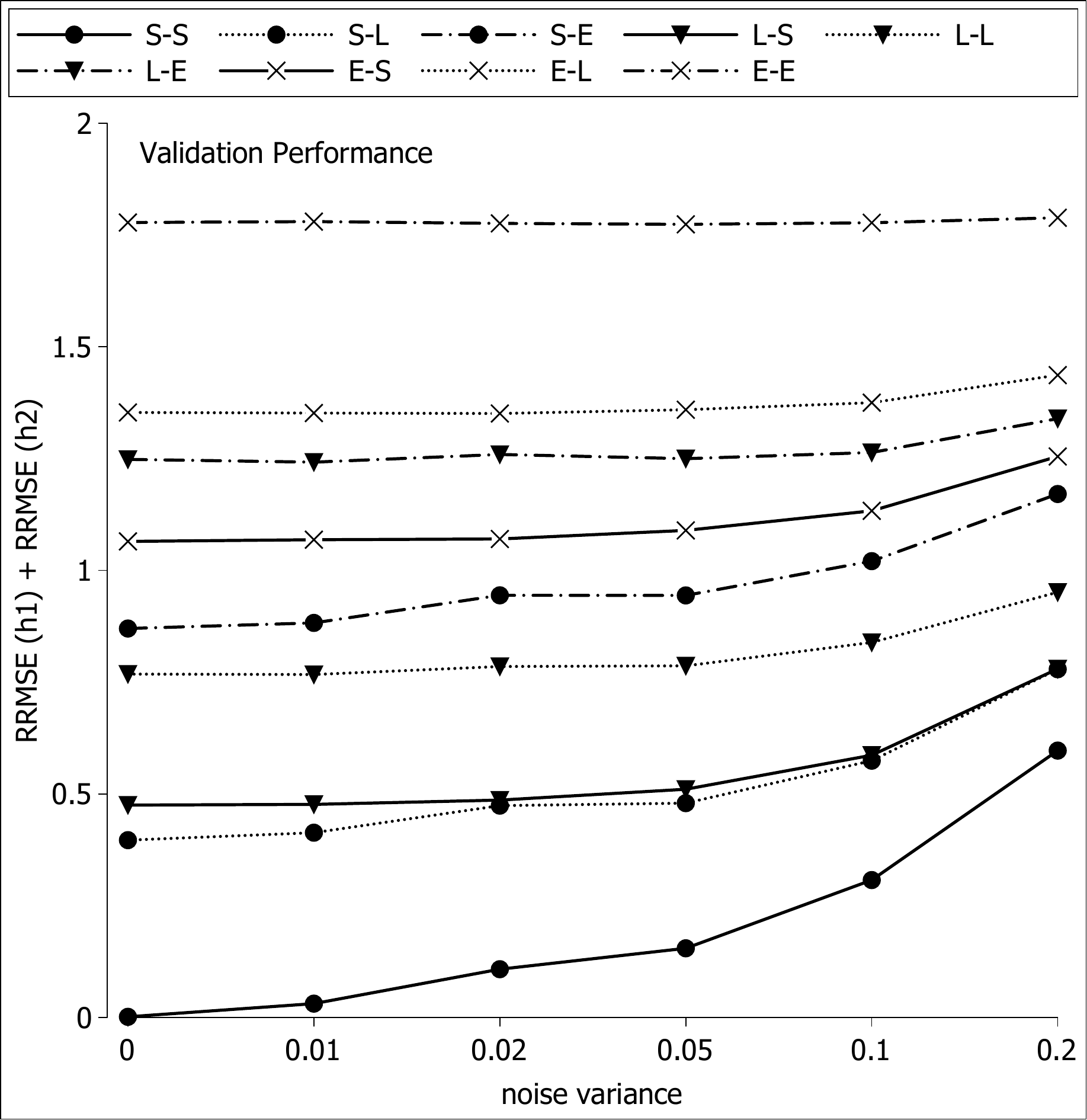}
			\subcaption{}
		\end{subfigure}
		\begin{subfigure}[t]{0.49\linewidth}
			\includegraphics[width=\linewidth]{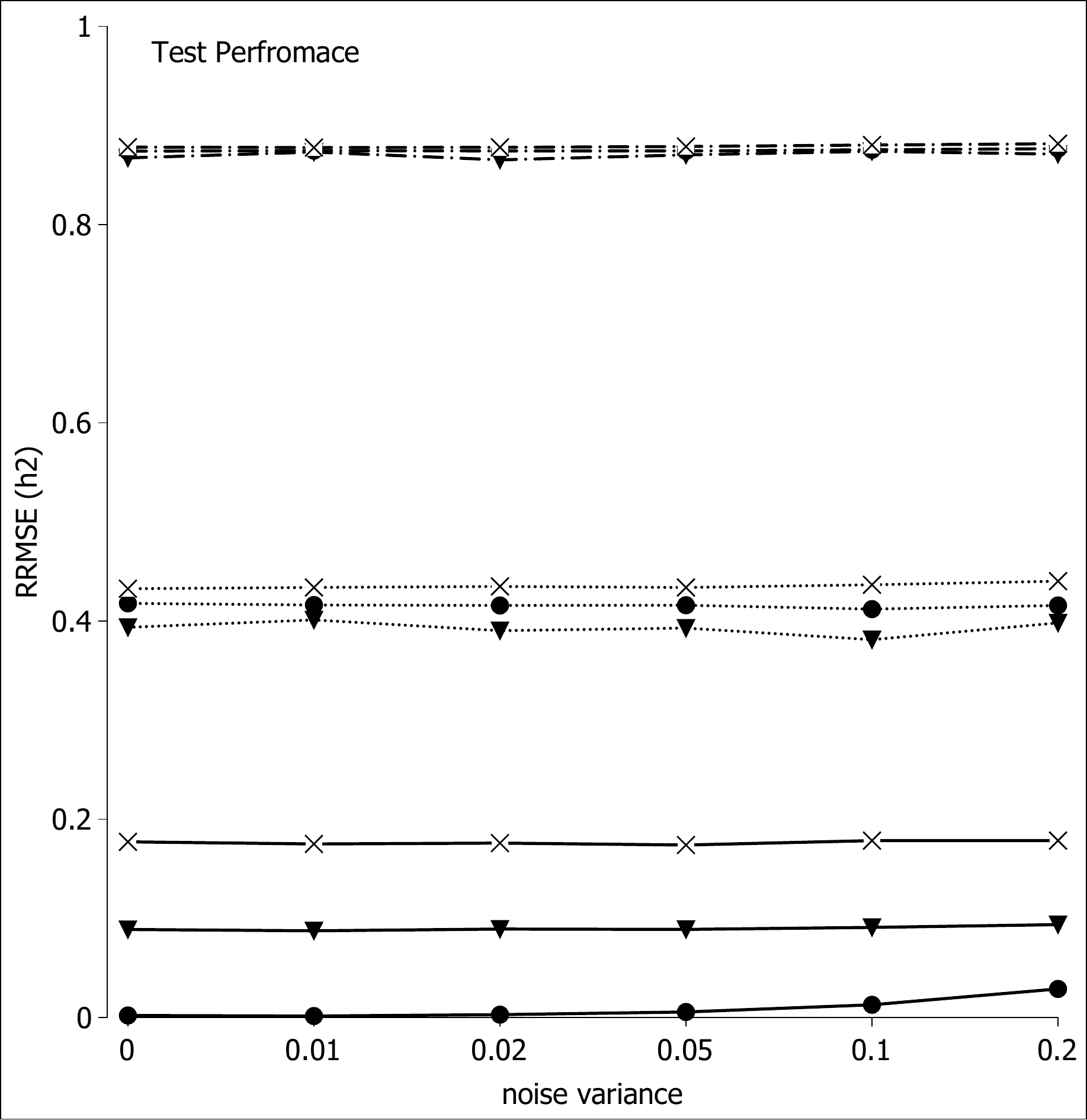}
			\subcaption{}
		\end{subfigure}
		\caption{\PBMhh{} : (a) Performance comparison of the summed validation errors of both signals $h_{1}$ and $h_{2}$  ($RRMSE_{h1}+RRMSE_{h2}$) from the 9 obtained models across 6 different synthetic cases. (b) Performance comparison test errors of the output signal $h_{2}$ ($RRMSE_{h2}$) from the 9 obtained models across 6 different synthetic cases. \PBMhh{} is able to accurately identify the correct structure based on the validation performance. The identified model exhibits robust and accurate predictive performance on test data.}
		\label{fig:ProbmotSTh1h2Noise}%
	\end{figure}

	While \twoPBM{} is able to correctly identify the ground-truth model, it follows a procedure that requires an intermediate intervention by the modeler. The decomposition and the ordering of stages is to be provided by a human expert that, in turn, has to run ProBMoT for each stage separately. In contrast, \PBMhh{} does not require any intervention by a modeler. Its identification capabilities are presented in Figure~\ref{fig:ProbmotSTh1h2Noise}. \PBMhh{} explores the complete space of nine candidate model structures, using both signals $h_{1}$ and $h_{2}$ as input, and ranks the models according to the summed error $RRMSE_{h1}+ RRMSE_{h2}$.
	
	\begin{figure}[!b]
		\centering
		\begin{subfigure}[t]{0.49\linewidth}
			\includegraphics[width=\linewidth]{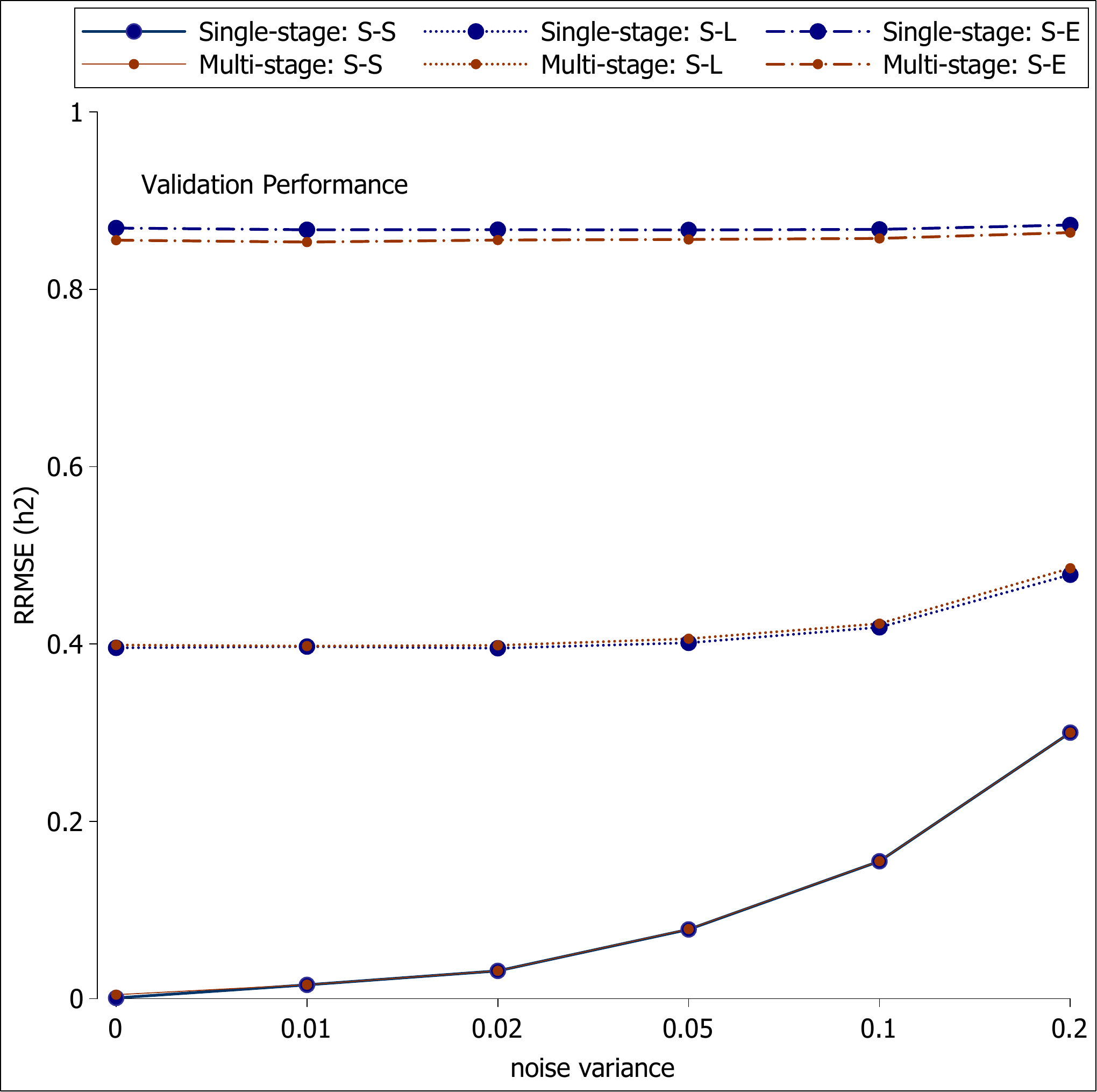}
			\subcaption{}
		\end{subfigure}
		\begin{subfigure}[t]{0.49\linewidth}
			\includegraphics[width=\linewidth]{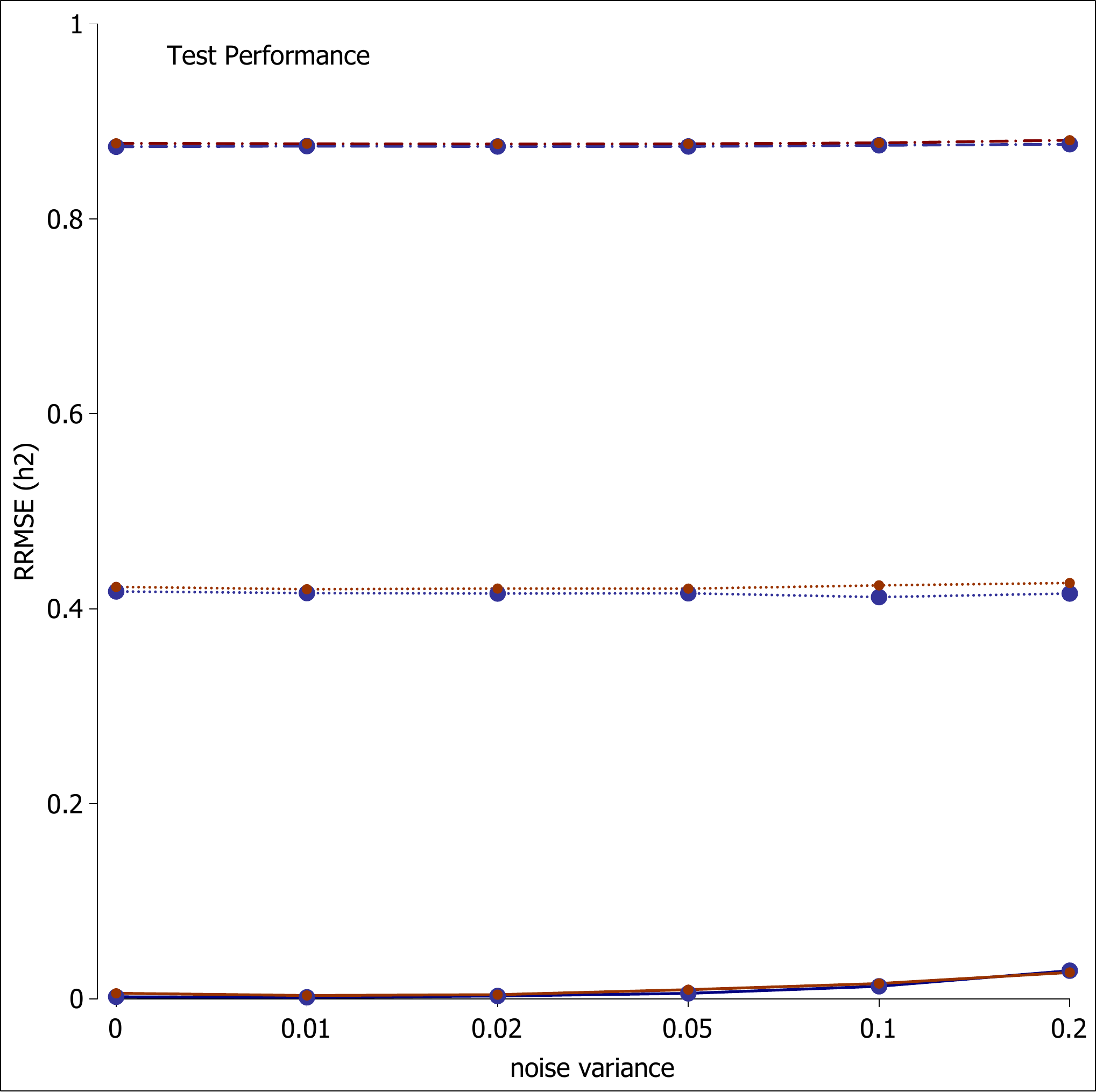}
			\subcaption{}
		\end{subfigure}
		\caption{ comparison of the errors of the three candidate models considered with \twoPBM{} and \PBMhh{}, measured on the response of the output signal $h_{2}$. Both modes accurately identify the correct structure of the model with virtually identical performance on both (a) validation and (b) test data.}
		\label{fig:StandardVS2ST}%

	\end{figure}
	
	\begin{figure*}[!t]
		
		\centering
		\begin{subfigure}[t]{0.24\textwidth}
			\includegraphics[width=0.95\textwidth]{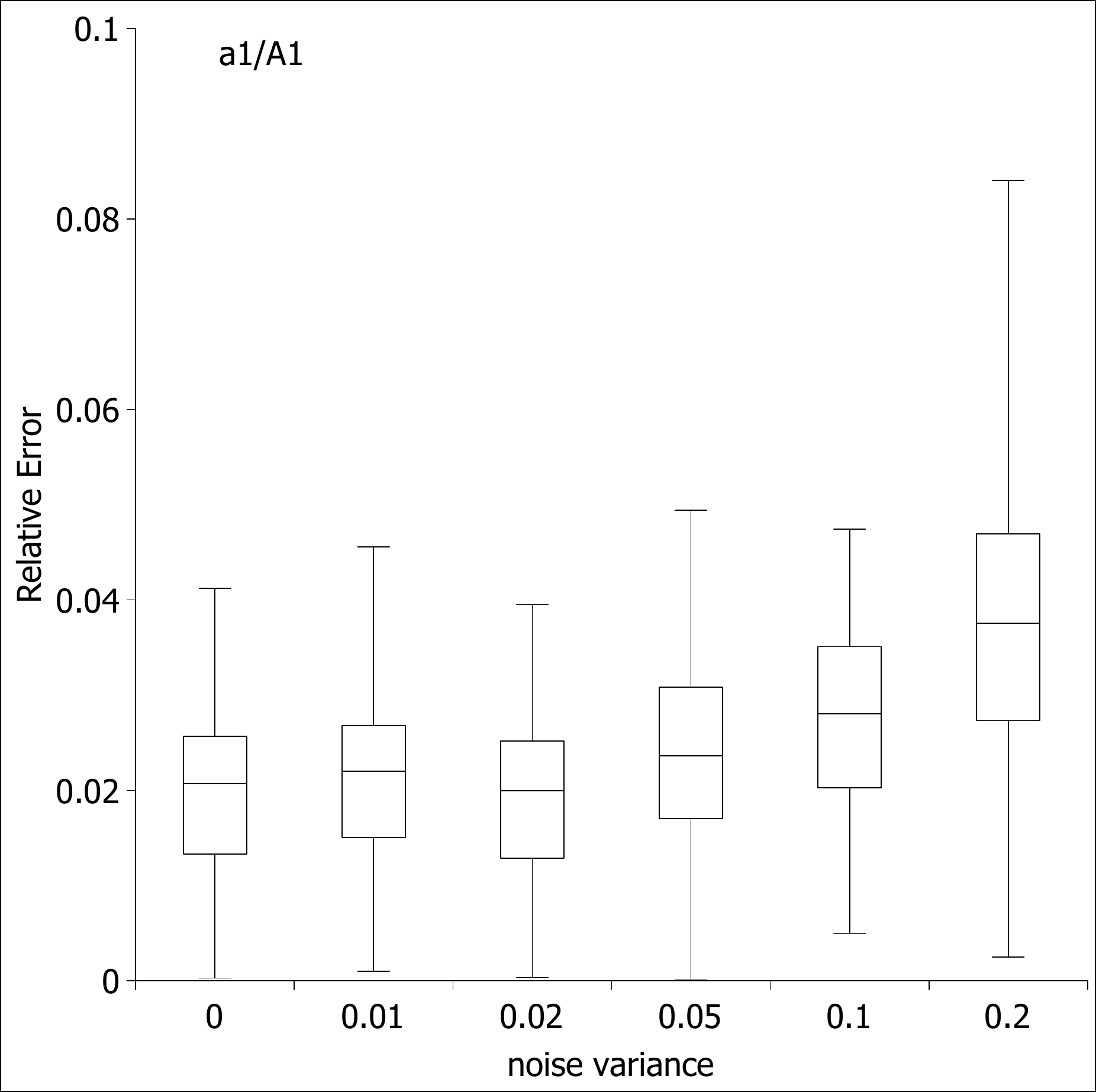}
			\subcaption{}
		\end{subfigure}
		\begin{subfigure}[t]{0.24\textwidth}
			\includegraphics[width=0.95\textwidth]{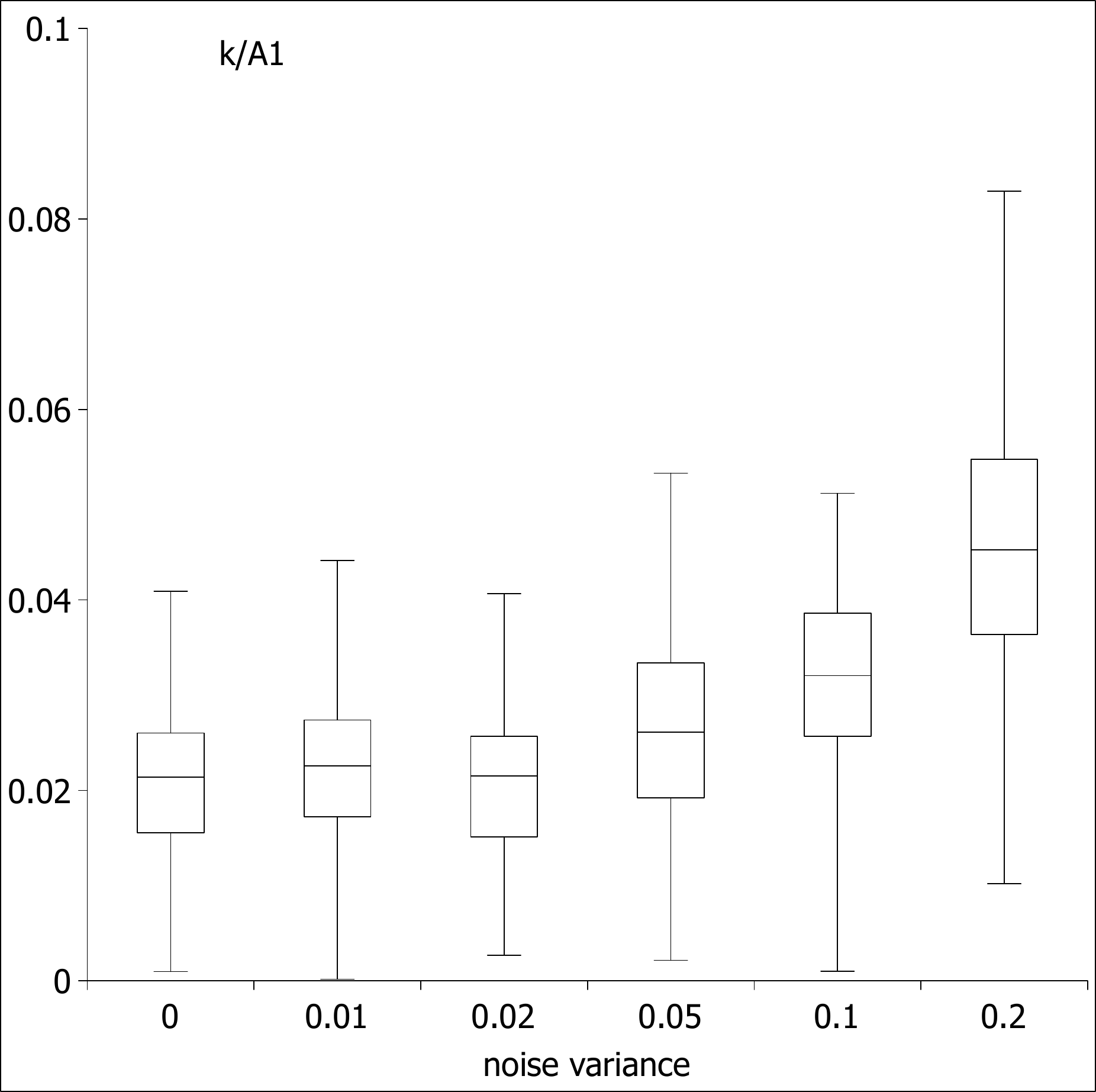}
			\subcaption{}
		\end{subfigure}
		\begin{subfigure}[t]{0.24\textwidth}
			\includegraphics[width=0.95\textwidth]{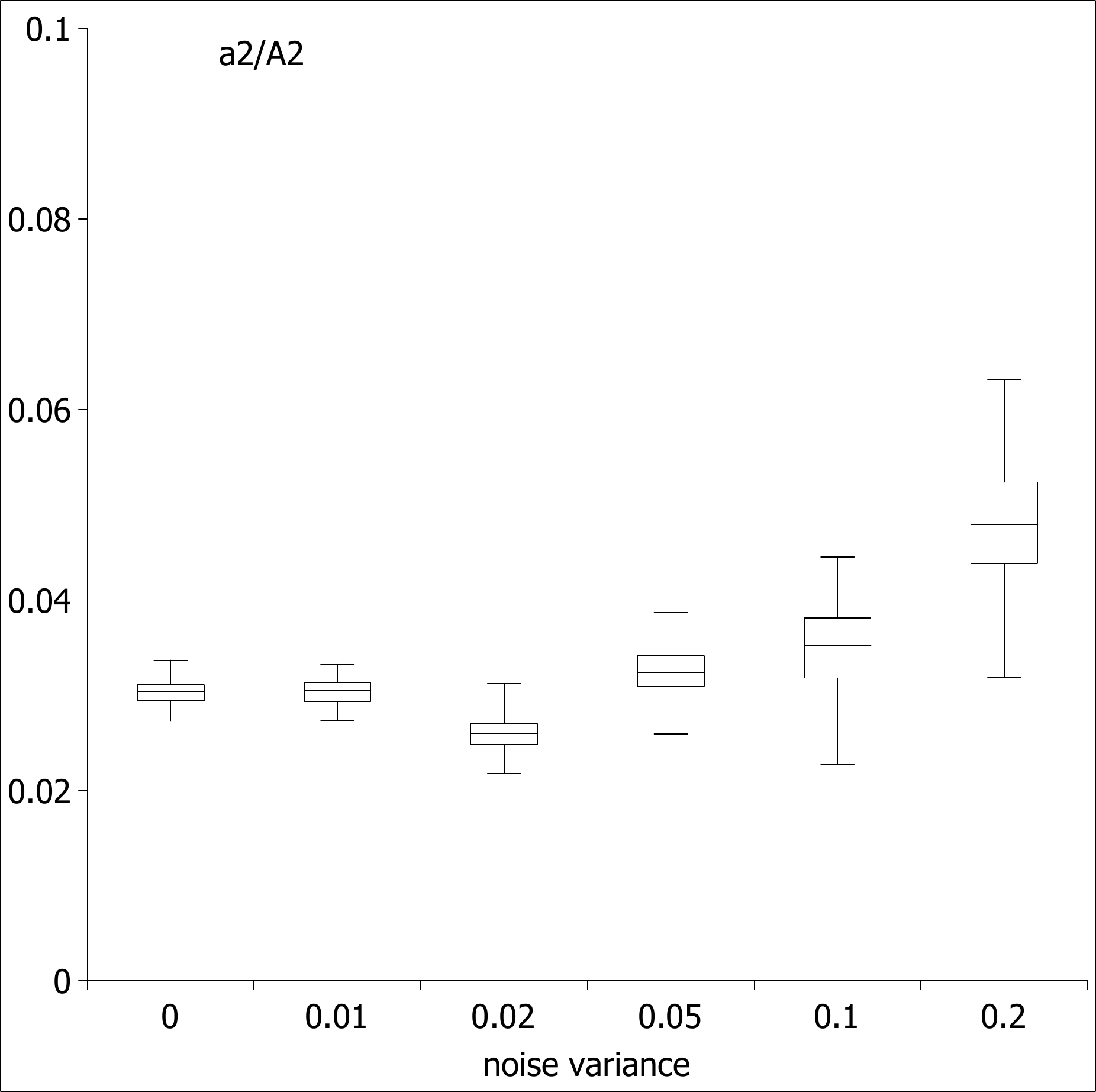}
			\subcaption{}
		\end{subfigure}
		\begin{subfigure}[t]{0.24\textwidth}
			\includegraphics[width=0.95\textwidth]{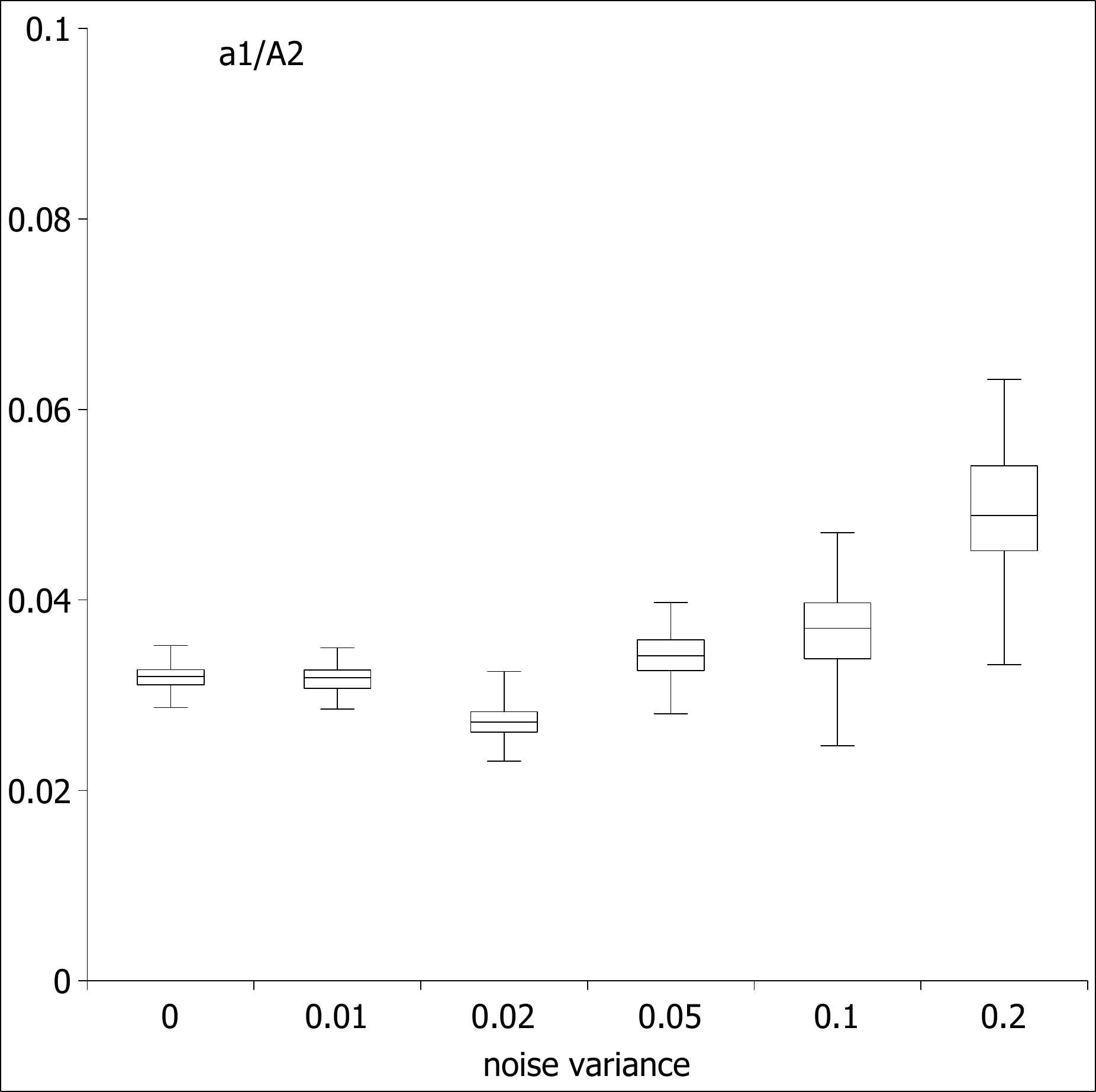}
			\subcaption{}
		\end{subfigure}

		\begin{subfigure}[t]{0.24\textwidth}
			\includegraphics[width=0.95\textwidth]{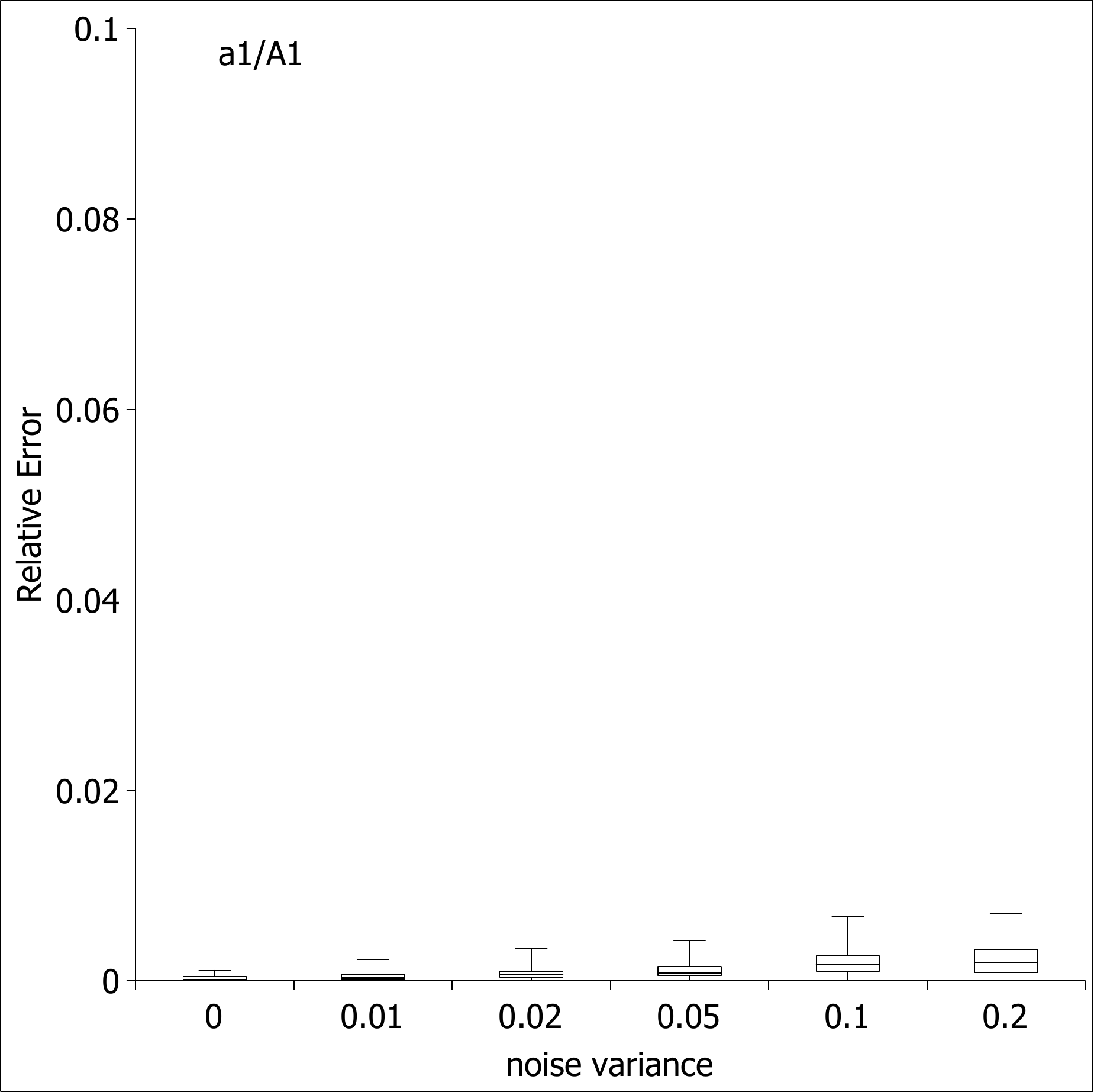}
			\subcaption{}
		\end{subfigure}
		\begin{subfigure}[t]{0.24\textwidth}
			\includegraphics[width=0.95\textwidth]{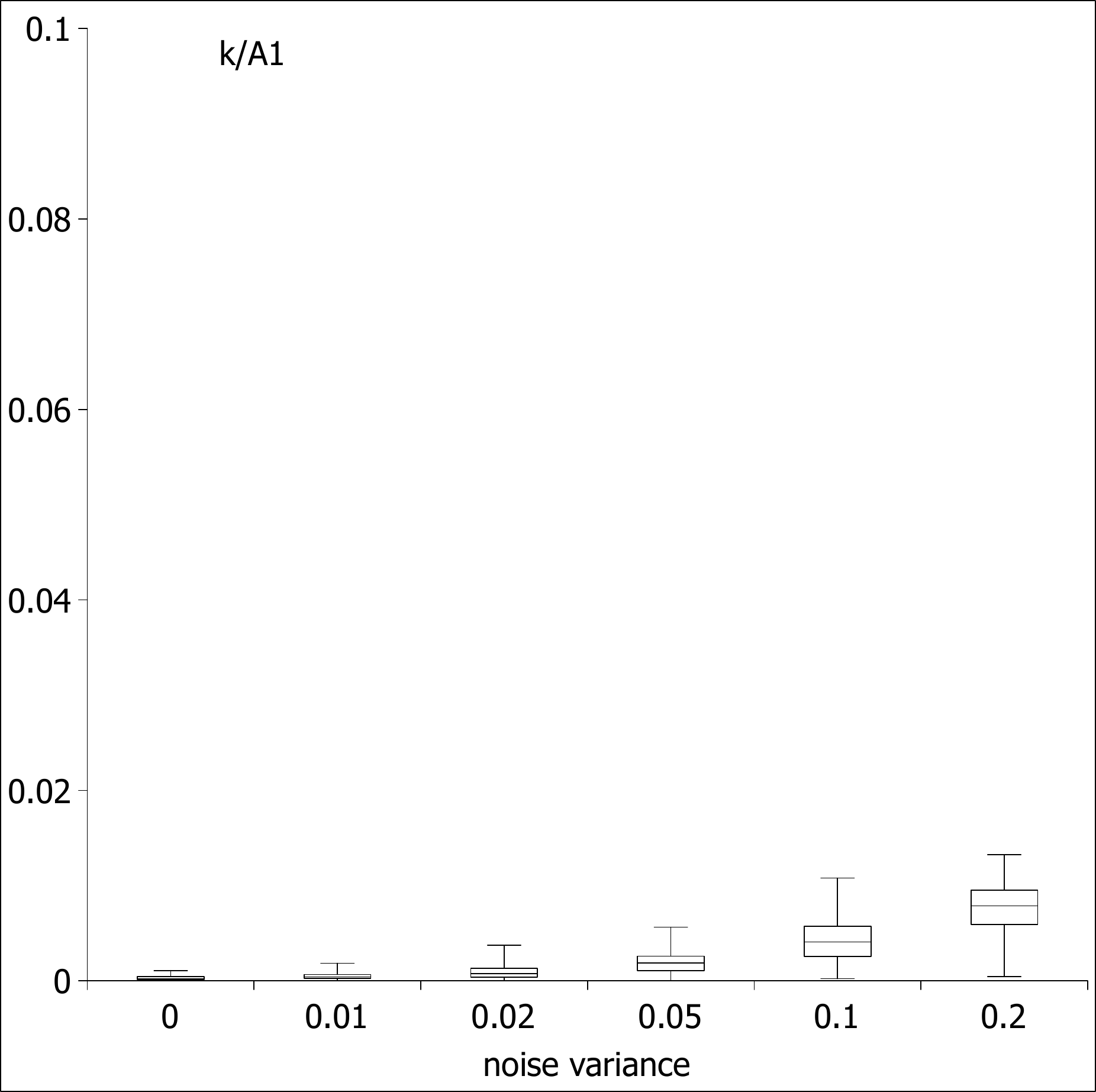}
			\subcaption{}
		\end{subfigure}
		\begin{subfigure}[t]{0.24\textwidth}
			\includegraphics[width=0.95\textwidth]{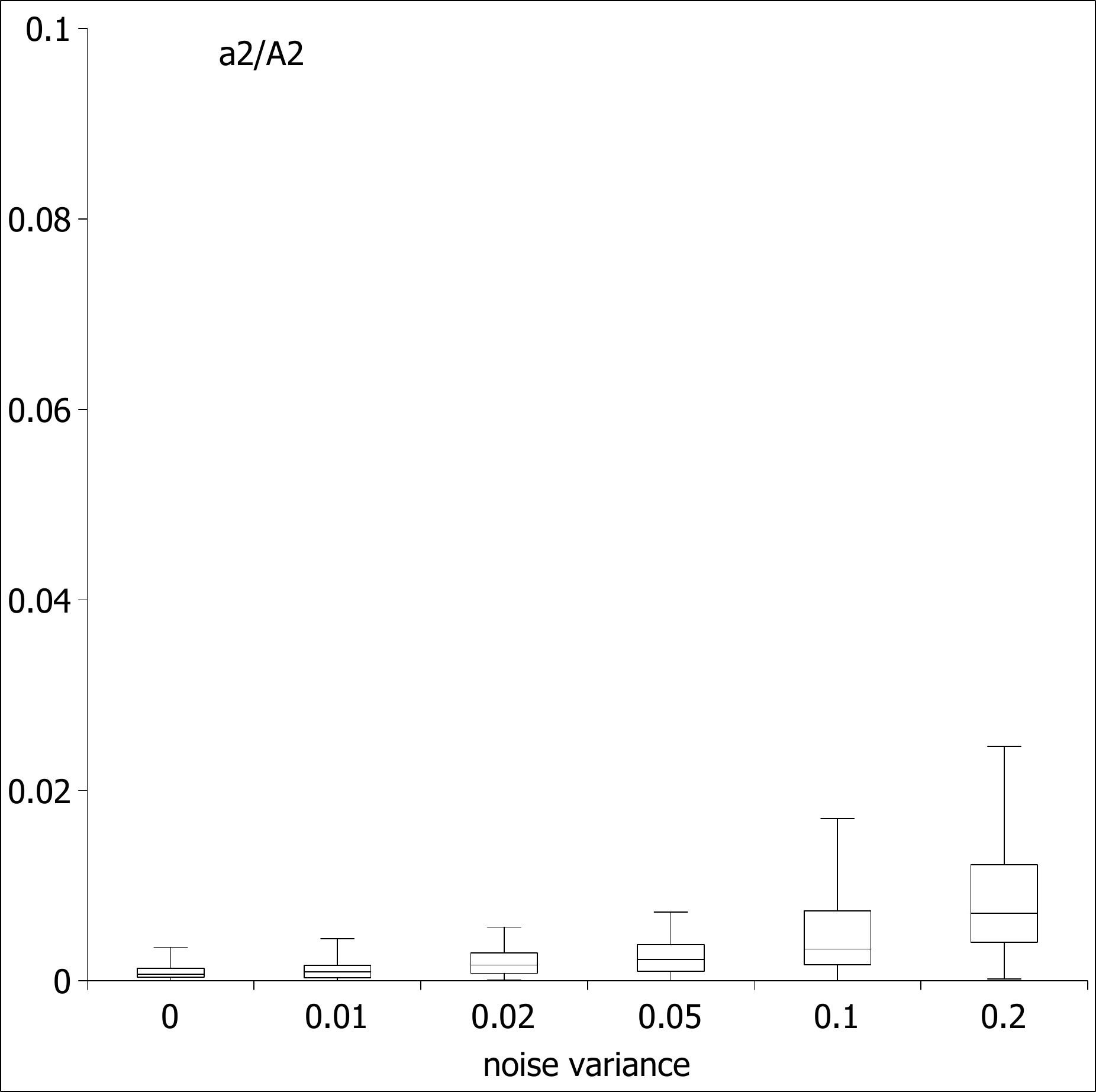}
			\subcaption{}
		\end{subfigure}
		\begin{subfigure}[t]{0.24\textwidth}
			\includegraphics[width=0.95\textwidth]{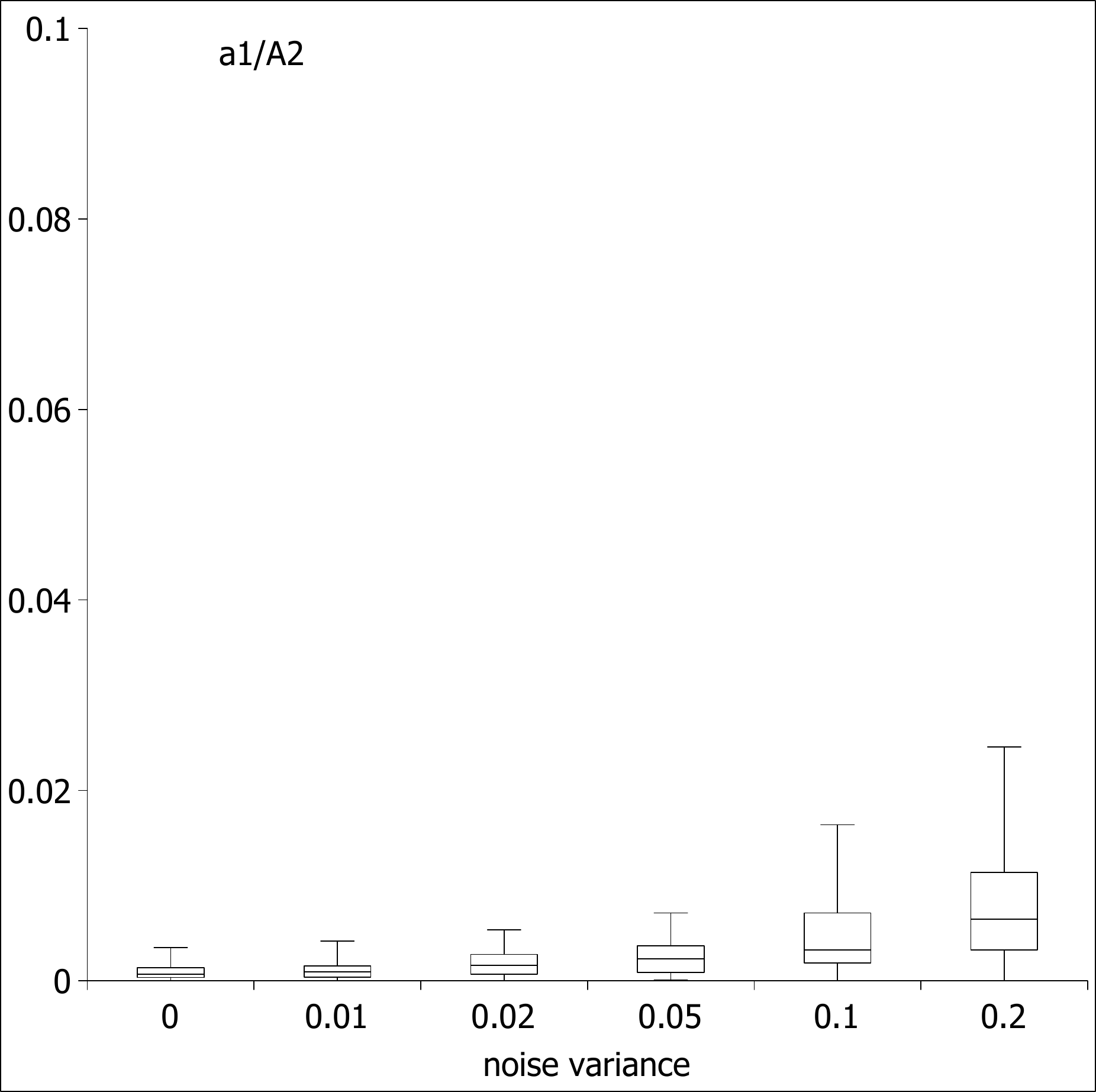}
			\subcaption{}
		\end{subfigure}
		
		\caption{Relative error between the ground-truth parameters and the fitted parameters estimated on various levels of noise using (\textbf{Top}) \twoPBM{} and (\textbf{Bottom}) \PBMhh{}. Here the relative errors are measured on the ratios of the 5 parameters : (a,e) $a_{1}/A_{1}$, (b,f) $k/A_{1}$, (c,g) $a_{2}/A_{2}$ and (d,h) $a_{1}/A_{2}$. While both \twoPBM{} and \PBMhh{} are able to accurately reconstruct the model parameters, \PBMhh{} exhibits stable behavior which results in relative difference between the fitted and the original parameter values that is lower than 1\%}
		\label{fig:Params}%
	\end{figure*}
	
	
	Figure~\ref{fig:ProbmotSTh1h2Noise}a shows that \PBMhh{} is able to reconstruct the correct model structure S-S: In particular, based on the performance on a separate validation set, \PBMhh{} accurately identifies the correct structure for all 6 synthetic cases. Note that here the relative difference in performance between the best identified structure and the remaining structures is substantial and higher than $20$ percentage points. Moreover, in terms of performance on unseen test data, \PBMhh{} exhibit stable behavior, where for all synthetic cases the discrepancy between the model's response of the output signal $h_{2}$ and the actual response (in the synthetic test data) remains stable and lower than $0.05$. 
	
	In sum, both modes of using ProBMoT accurately identify the correct structure of the model. Figure~\ref{fig:StandardVS2ST} presents a comparison of the errors of the three candidate models considered with \twoPBM{} and \PBMhh{}, measured on the output signal $h_{2}$. Figure~\ref{fig:StandardVS2ST} shows that both methods lead to models with virtually identical errors on both validation (Figure~\ref{fig:StandardVS2ST}a) and test (Figure~\ref{fig:StandardVS2ST}b) data. This means that for the task of structure identification, \PBMhh{} without any intervention by a modeler, explores the complete (and larger) space of candidate model structures, performs almost identically to \twoPBM{}. The latter embodies the state-of-the art approach in system identification, where the modeling task is tackled via decomposition into multiple stages, and requires intervention by a modeler.
	
	Finally, we assess the capability of both \twoPBM{} and \PBMhh{} to reconstruct the values of the model parameters from noisy data. Figure~\ref{fig:Params} depicts the relative difference between the estimated parameter values in the learned models and the original parameter values used for generating the synthetic data. To obtain reliable parameter estimates, we performed 100 repetitions of parameter estimation with Differential Evolution (by varying the random seed) for each of the six synthetic cases. Note that, while the model structure has 5 parameters, in Fig~\ref{fig:Params}, we present the relative errors for the constants that appear in the equation from Fig~\ref{tab:pbmlib-wt}c, i.e. the ratios of the parameters $a_{1}/A_{1}$, $k/A_{1}$, $a_{2}/A_{2}$ and $a_{1}/A_{2}$, respectively.
	
	As we can see from Fig~\ref{fig:Params}, both modes of using ProBMoT can accurately estimate the original parameters values regardless of the noise variance. In general, the relative difference between the estimated and the original parameter values for both methods is lower than 5\% across all experimental cases. Note, however, that \PBMhh{} performs much better, lowering the variance of the parameter estimates to below 1\%. However, for the particular modeling task at hand, the differences in the parameter values obtained with the two modes of ProBMoT is not substantial, since both sets of parameter values can accurately model the dynamics of the water-tanks system.

	\subsection{Results on measured data}
	
	Here we focus on testing the ability of the process-based modeling approach to accurately reconstruct the benchmark system from measured data. Note that, for this modeling attempt, we assume that the correct structure of the system corresponds to the one given in Figure~\ref{fig:pbm-wt}c while the parameters are unknown and remain to be found. More specifically, since both \twoPBM{} and \PBMhh{} performed identically on the synthetic experiments, here we only use \PBMhh{} to correctly recover the structure of the model and estimate the models parameters. Table~\ref{tab:PBMReal} presents the results of this experiment.
	
	\begin{table}[b!]
	\centering
		\small
		\caption{The summed validation error of both signals $h_{1}$ and $h_{2}$  ($RRMSE_{h1}+RRMSE_{h2}$) and the test error of the output signal $h_{2}$ ($RRMSE_{h2}$) from the 9 models obtained from measured data.}
		\begin{tabular}{l|r|r}
				\hline
				\multirow{2}[2]{*}{Model} & \multicolumn{1}{c|}{$RRMSE_{h1}+RRMSE_{h2}$} & \multicolumn{1}{c}{$RRMSE_{h2}$} \bigstrut[t]\\
				& \multicolumn{1}{r|}{VALIDATION} & \multicolumn{1}{r}{TEST} \bigstrut[b]\\
				\hline
				S-S     & \textit{0.481} & \textbf{0.23} \bigstrut[t]\\
				S-L     & 0.539   & 0.306 \\
				S-E     & 1.038   & 0.805 \\
				L-S     & \textbf{0.44} & \textit{0.249} \\
				L-L     & 0.494   & 0.313 \\
				L-E     & 0.999   & 0.805 \\
				E-S     & 0.977   & 0.248 \\
				E-L     & 1.125   & 0.37 \\
				E-E     & 1.592   & 0.818 \bigstrut[b]\\
				\hline
			\end{tabular}%
				\label{tab:PBMReal}%
	\end{table}%
	
		\begin{figure}[b!]{}
			\centering
			\includegraphics[width=\linewidth]{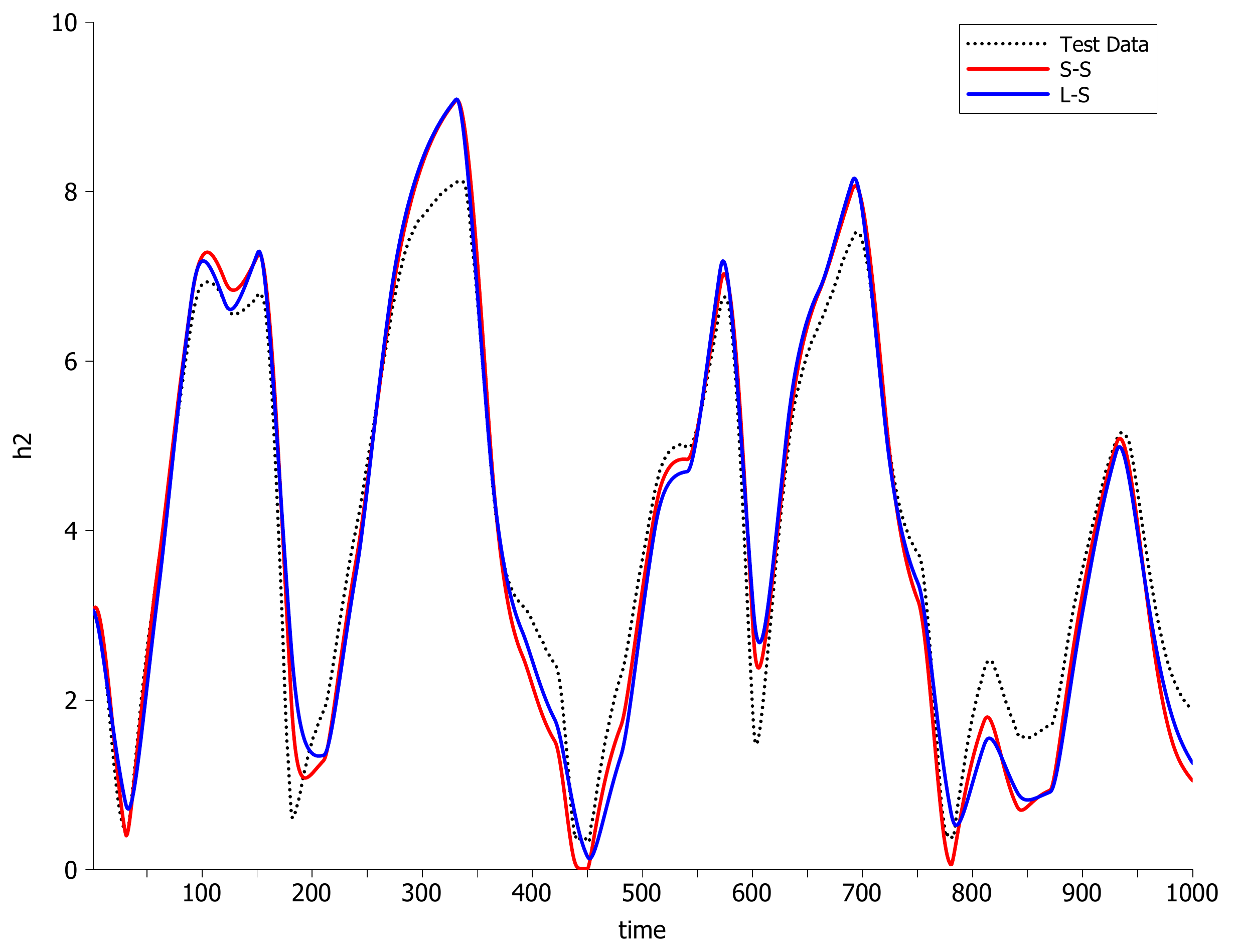}
			\caption{Simulation responses on the test data of the two top-ranked models S-S and L-S obtained from measured data. The simulated responses of both models are almost identical.}
			\label{fig:PBMReal}
		\end{figure}

	From Table~\ref{tab:PBMReal} we can see that \PBMhh{} struggles to identify the assumed correct structure from the measured data. Namely, we can see that the L-S model performs better than the S-S model in terms of summed $RRMSE$ on both signals. Nevertheless, in terms of test performance, the S-S model has the best performance on the test set with the model L-S being second. Note however, that the difference in test performance between S-S and L-S remains low and insignificant in favor of the model S-S. To properly assess these behaviors, Fig~\ref{fig:PBMReal} presents the simulation responses on the test data of the S-S and L-S models. From the figure, we can see that the simulated responses are indistinguishable. We conjecture that this is due to an implicit noise in the data i.e. from unforeseen nonlinearities in the measuring set-up, which have not been included in the knowledge provided at input.

	\subsection {Discussion}
	
	In general, the results from the empirical analyses show that process-based modeling can be successfully applied to the task of nonlinear systems identification. In particular, from the experiments performed in the synthetic cases, we can conclude that ProBMoT, using domain knowledge and data, can accurately identify the correct structure of the system as well as estimate the model's parameter values with high accuracy, regardless of the signal corruption present in the data. In the measured-data experiments, however, ProBMoT performs moderately well by discarding the majority of the models with incorrect structures, but nevertheless struggling to identify the "correct" one among the top two ranked. Earlier, we conjectured that such a performance might be an artifact from some nonlinearities in the measuring set-up which were not captured in the domain knowledge provided at input. Such lack of knowledge can be explicitly handled by the modeler, where one can additionally decide on the correct structure. However, in the case of process-based modeling, such behavior can also be handled by extending the modeling knowledge provided at input: Either by encoding new processes in the library of domain knowledge or by weakening some of the constraints that relate to the dynamics of the individual processes.


	For this particular task, we extend the modeling knowledge by weakening the constraint on the two processes {\tt\sc ValveTransmission} and {\tt\sc Outflow}. Given that ProBMoT mostly struggled to differentiate between strictly linear and nonlinear (square-root) components for the interactions that involve the two tanks $h_{1}$ and $h_{2}$, we add a component that allows for all sub-linear interactions. We encode this in the process-based library Table~\ref{tab:LibPower}, by adding a new process alternative {\tt\sc Power} to each of the processes {\tt\sc ValveTransmission} and  {\tt\sc Ouflow}. These two alternatives in turn are instantiated to two additional components which allow for modeling any sub-linear dynamics in {\tt\sc ValveTransmission} and {\tt\sc Ouflow}. In particular, both processes introduce two additional parameters $P_{h1}$ and $P_{h2}$,  where:
	\begin{equation*}
	0 < P_{h1}, P_{h2} < 1 ,
	\end{equation*}
	
	which correspond to the exponents of the two signals $h_{1}$ and $h_{2}$, respectively, in the equations. Table~\ref{tab:POWresults} presents the results of the modeling experiment given the extended process-based library. To test this extension, we perform experiments using measured data, where we evaluate the performance of a new model P-P and compare its performance against the models S-S and L-S.

	
	\begin{table}[]
		\caption{Extension of the library for the cascaded water tank system with new process alternatives.}
		\centering
		\ttfamily 
		\begin{tabular}{l}
			\hline
			\vdots \\
			
			\color{RoyalBlue}template process \textsc{Power}:\texttt{\textsc{ValveTransmission}} \{ \\
			
			\color{RoyalBlue}	\qquad \texttt{equations} : \\
			\color{RoyalBlue}	\qquad\qquad	$\begin{aligned}  & \mathrm{td(t1.h) = - G \cdot pow(t1.h, \textbf{P})\cdot t1.a/t1.A},\\ 
			& \mathrm{td(t2.h) = G \cdot pow(t1.h, \textbf{P})\cdot t1.a/t2.A}; \}\\
			\end{aligned}$\\
			\vdots\\
			
			\color{Green}	template process \textsc{Power}:\texttt{\textsc{Outflow}} \{ \\
			\color{Green}\qquad 	equations: \\ 
			\color{Green}	\qquad	\qquad $\mathrm{td(t.h) = - G \cdot pow(t.h, \textbf{P}) \cdot t.a/t.A }$;
			\} \\ 
			\vdots\\
			
			\hline
			
		\end{tabular}%
		\label{tab:LibPower}%
	\end{table}%
	
\begin{figure}[]
			\centering
			\includegraphics[width=\linewidth]{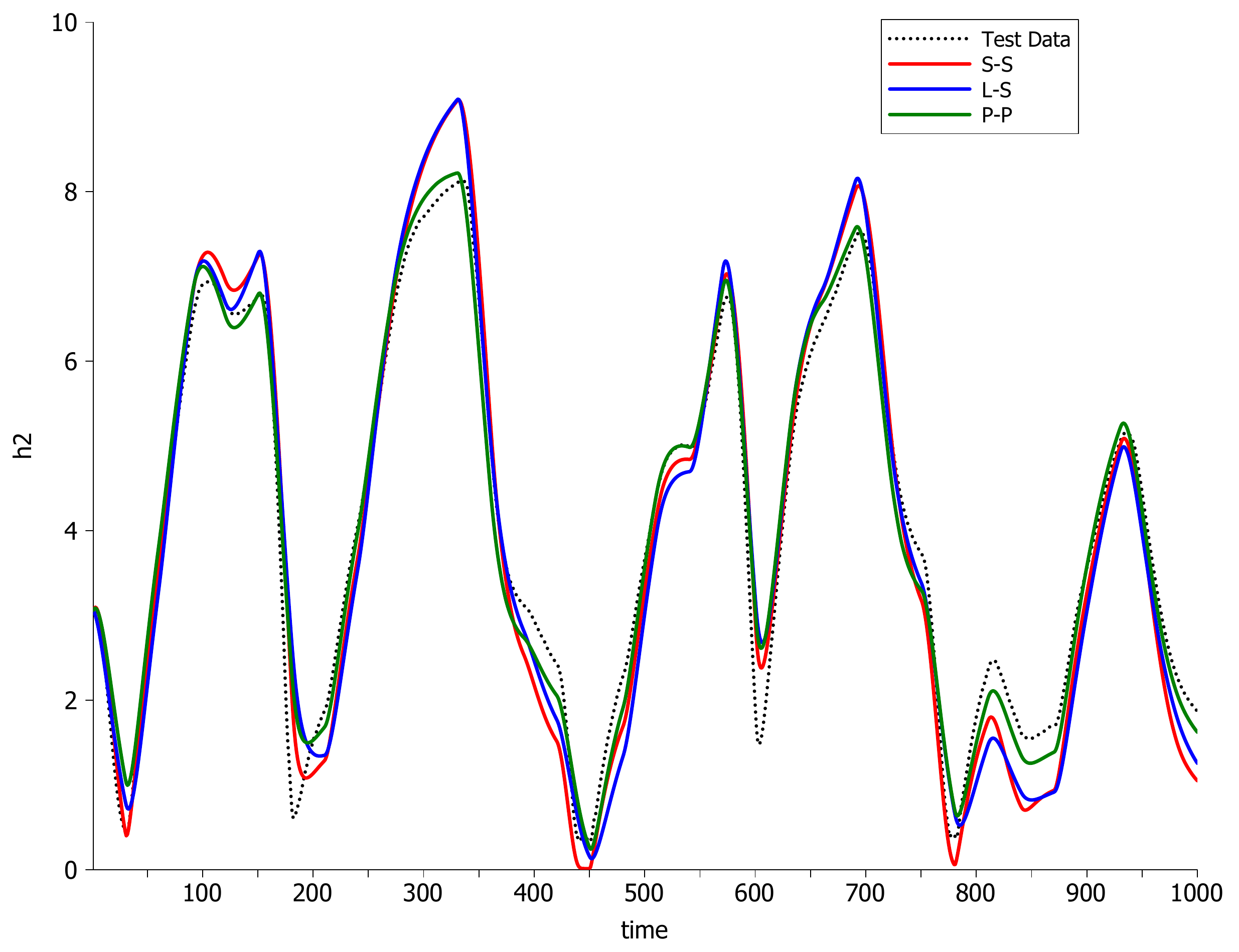}
			\caption{Simulation responses on the test data of the constructed models P-P, S-S and L-S obtained from measured data. The simulated response of the P-P model better captures the measured behaviour than the remaining two models.}
			\label{fig:POWresults}%
		\end{figure}

	The results reported in Table~\ref{tab:POWresults} show that the model P-P performs substantially better on the validation set ($RRMSE_{h1}+RRMSE_{h2}$) than both S-S and L-S models. This result confirms our hypothesis that, for this particular modeling task, extending the modeling knowledge allows for better identification. Note that the estimated parameter values of  $P_{h1}$ and $P_{h2}$ for the signals $h_{1}$ and $h_{2}$ are $0.68$ and $0.66$, respectively. This results is also in-line with our hypothesis that there are some implicit sub-linear dynamics captured in the measured data, which does not correspond to the original theoretical model from Figure~\ref{fig:pbm-wt}c. Moreover, Figure~\ref{fig:POWresults} presents the simulated responses on the test set of the three models P-P, S-S and L-S. From the figure, we can see that, for some of the trends where S-S and L-S struggle, P-P is able to capture them with high accuracy. In turn, this accuracy translates to predictive error on the output signal $h_{2}$ of $RRMSE_{h2}= 0.164$ in the case of P-P, $RRMSE_{h2}= 0.23$ in the case of S-S and $RRMSE_{h2}= 0.249$ of the model L-S.  
	
	\begin{table}[b!]
	\centering
		\small
		\caption{The summed validation errors of both signals $h_{1}$ and $h_{2}$ ($RRMSE_{h1}+RRMSE_{h2}$) and test errors of the output signal $h_{2}$ ($RRMSE_{h2}$) from the new established model P-P and the two models S-S and L-S constructed from measured data.}
	
			\begin{tabular}{l|r|r}
				\hline
				\multirow{2}[2]{*}{Model} & \multicolumn{1}{c|}{$RRMSE_{h1}+RRMSE_{h2}$} & \multicolumn{1}{c}{$RRMSE_{h2}$} \bigstrut[t]\\
				& \multicolumn{1}{r|}{VALIDATION} & \multicolumn{1}{r}{TEST} \bigstrut[b]\\
				\hline
				S-S     & 0.481 & 0.23 \bigstrut[t]\\
				L-S     & 0.44 & 0.249 \\
				P-P   &\textbf{0.238} & \textbf{0.164} \bigstrut[b]\\
				\hline
			\end{tabular}%
			\label{tab:POWresults}%
	\end{table}
	
\begin{figure}[t!]
		\centering
		\begin{subfigure}[t]{0.49\linewidth}
			\includegraphics[width=\linewidth]{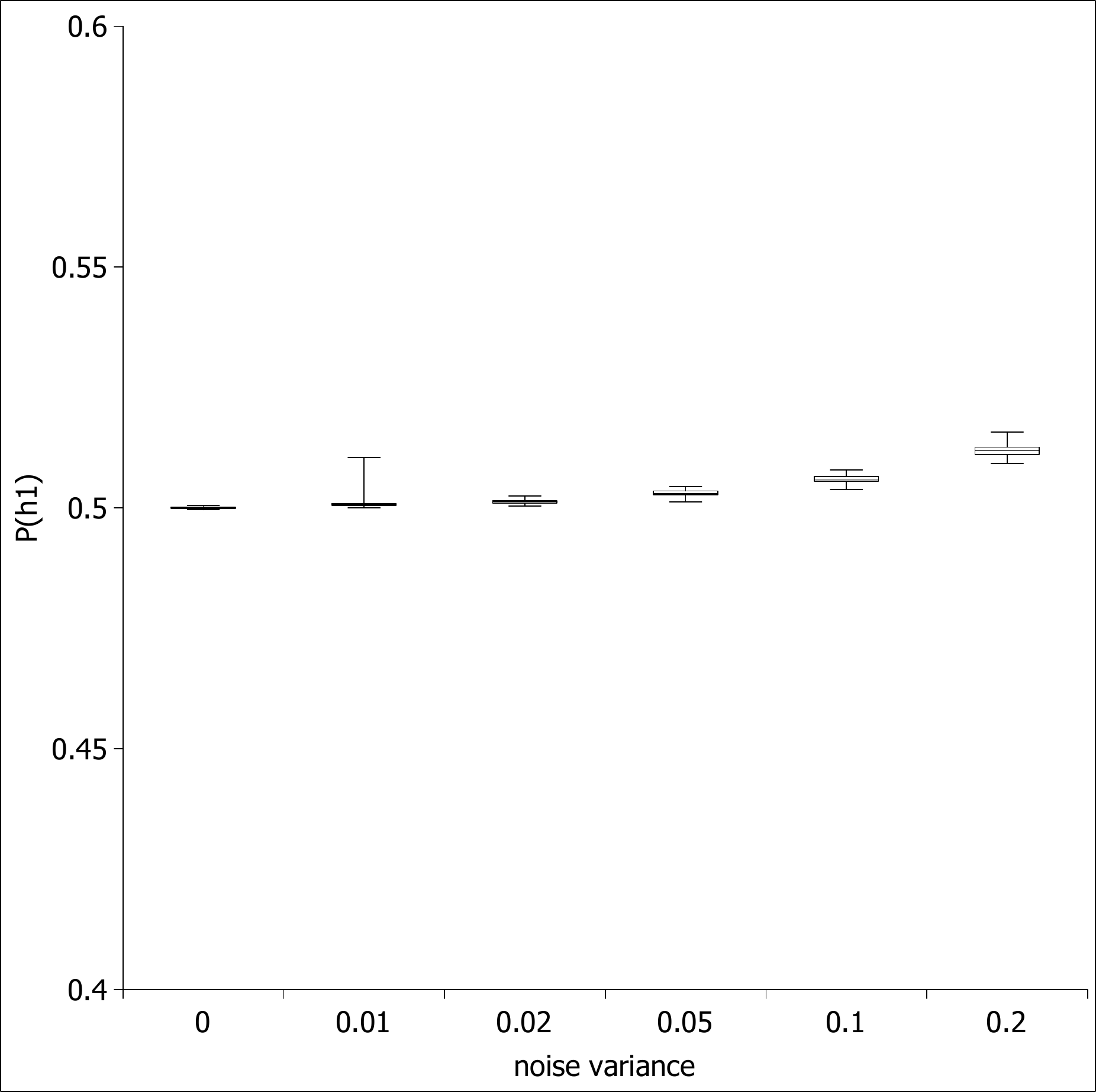}
		\end{subfigure}
		\begin{subfigure}[t]{0.49\linewidth}
			\includegraphics[width=\linewidth]{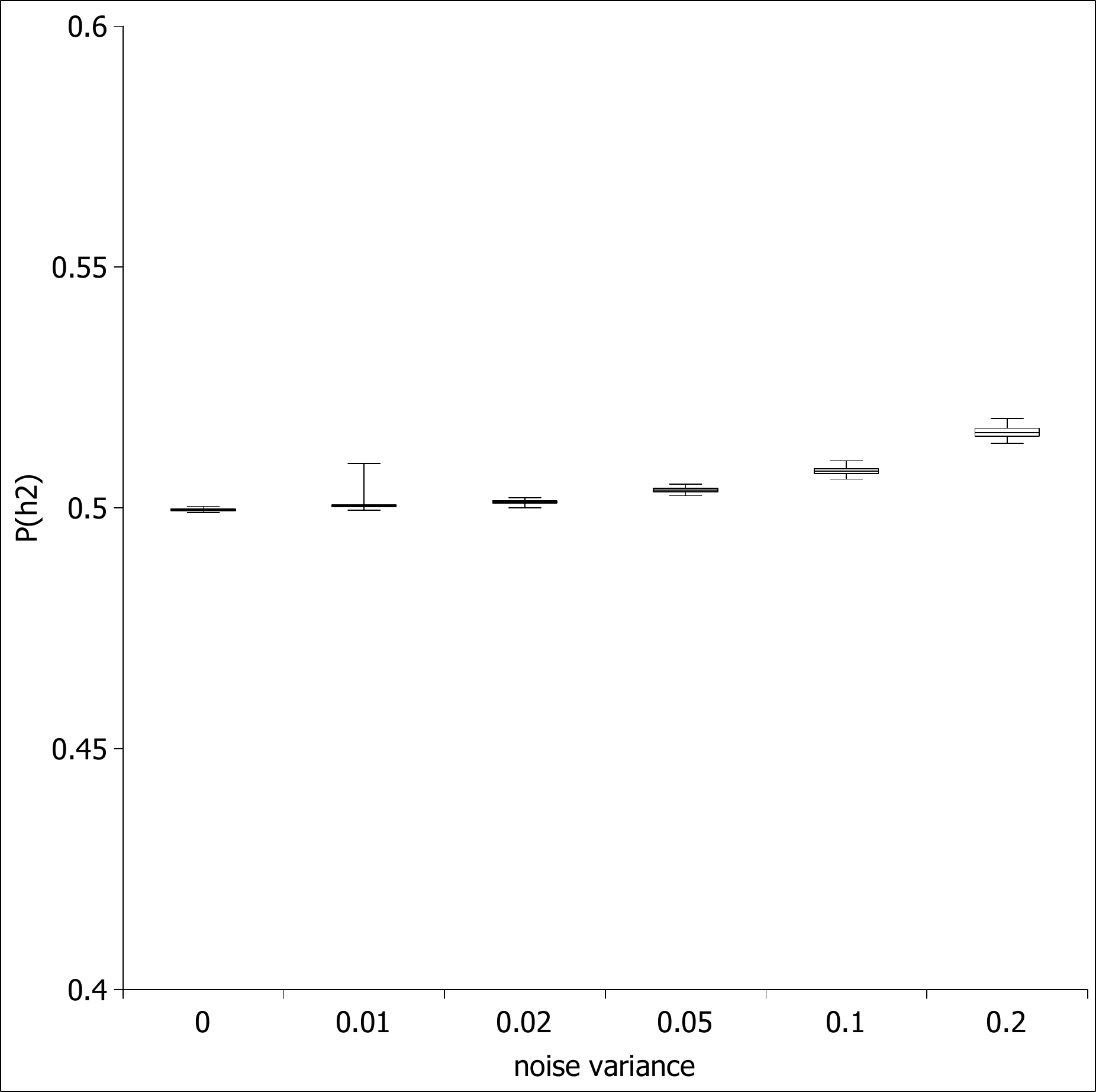}
		\end{subfigure}
		\caption{Estimated values of the two parameters $P_{h1}$ and $P_{h2}$ across the six synthetic data sets. ProBMoT is able to accurately estimate the values of both parameters in the range of $0.5$ which yield an identical model structure to the correct one i.e. S-S. }
		\label{fig:POW_parameter}%
	\end{figure}

	In sum, we showed that by weakening the constraints of the processes {\tt\sc ValveTransmission} and {\tt\sc Outflow} in the process-based library, ProBMoT is able to discriminate among the model structures P-P, on one hand, and the remaining two, on the other, therefore more accurately modeling the benchmark system from measured data. Note that, while we implemented this extension by adding two additional parameters, the performance improvement is a direct result of changing the structure of the model. To test this hypothesis, we performed an additional parameter estimation experiment for estimating the values of $P_{h1}$ and $P_{h2}$ using the synthetic data sets. For this task, we performed 100 repetitions for each of the 6 different synthetic cases. The results for both parameters are given in Figure~\ref{fig:POW_parameter}.

	
	From Figure~\ref{fig:POW_parameter} we can see that ProBMoT is estimates the values of both parameters to be in the range from $0.5$ (with standard deviation $\sigma=10^{-4}$) in the synthetic case with noise variance $0$ to $0.51$ (with standard deviation $\sigma=10^{-3}$) in the case with noise variance $0.2$. This stable behavior means that, in a controlled scenario where the appropriate amount of knowledge is present in the process-based library, ProBMoT is able to identify a structure identical to the underlying correct model structure (S-S) regardless of the corrupted responses of the input/output signals.
	

	\section{Conclusion}
	

	In this paper, we present the first proper evaluation of process-based modeling in the context of system identification. Process-based modeling, a recent equitation discovery approach, addresses the task of computer-aided modeling of nonlinear dynamic systems from measurements and domain-specific modeling knowledge. To test its utility, we perform a systematic evaluation of its utility for reconstructing a standard system identification benchmark model of cascaded water-tank system. To this end, we test the latest implementation of process-based modeling, i.e. the ProBMoT system, which allows for computer-aided construction, long-term simulation and evaluation of process-based models of dynamic systems. More specifically, we first analyze the robustness of ProBMoT on a series of identification tasks involving synthetic data with different noise variances, obtained by simulating the benchmark model. Next, we investigate ProBMoT's ability to establish a valid model of a real benchmark system from measurement data.
	
	The empirical evaluation shows that ProBMoT can reconstruct both the structure and the parameters of a system from synthetic data with various levels of corruption, as well as measured data. In particular, in the synthetic-data experiments, ProBMoT is able to accurately identify the correct structure of the system, as well as to estimate the model's parameter values with high accuracy, regardless of the signal corruption present in the data. On the other hand, in the measured-data experiments, where ProBMoT initially struggled to discriminate between the two top-ranked models, we were able to identify implicit nonlinearities that were present in the data and not in the modeling knowledge provided at input. Consequently, with a minimal extension of the process-based library, we showed that ProBMoT can successfully reconstruct both the structure and the parameters of the model from measured data.
	
	The process-based modeling approach, and in particular its application to the task of nonlinear system identification, is the main contribution of this paper. Namely, PBM has several distinguishing properties. First, PBM is a "white-off" gray-box modeling approach \cite{Ljung2008} that constructs models from components with a clear mapping to the modeled physical phenomena. Next, the resulting process-based models retain the utility of mathematical models and therefore can be simulated and analyzed using well established numerical approaches. Finally, PBM is general and can be applied to any domain and to any other task of system identification. Moreover, process-based knowledge employed by the PBM is domain-specific in the sense that it captures the basic modeling principles in a given domain and can be reused for different modeling applications within the same domain.
	
	Several directions for further work can be followed. While our empirical study is limited to one domain, i.e. modeling the dynamics of a cascaded water tank system with two tanks, the proposed approach to learning process-based models can be applied to many other tasks of system identification. An immediate continuation of the work presented here would be to investigate whether the findings of this paper also apply to other standard nonlinear benchmark systems in both real-world and synthetic settings. Next, we intend to compare the process-based modeling approach to other state-of-the-art gray-box modeling approaches, both in terms of accuracy of reconstructing the dynamics of a nonlinear system and in terms of the comprehensibility of the models that these approaches can provide. 
	

\section*{Acknowledgment}
		We would like to acknowledge the financial support of the Slovenian Research Agency, via the grants P2-0103, P2-0001, P5-0093 and N2-0056.


%

  \newpage

\appendices
	\section*{Appendix A: Modeling Scenarios}
	\begin{table}[h]
		\caption{Modeling scenario with \PBMhh{}. The initial values of $h_{1}$ and $h_{2}$ are taken from the measured data.}
		\centering
		\ttfamily 
		\begin{tabular}{>{\footnotesize  }l}
			\hline
			entity \textsl{tank1} : \textsl{Tank} \{ \\
			\qquad vars: \textsl{h} \{role: \textsl{endogenous}; initial: 0.38086;\};\\
			$\begin{aligned}
			\qquad \texttt{consts} : \quad  &\textsl{A}= \texttt{null}, \\
			&\textsl{a} = \texttt{null}; \}\\
			\end{aligned}$\\
			entity \textsl{tank2} : \textsl{Tank} \{ \\
			\qquad vars: \textsl{h} \{role: \textsl{endogenous}; initial: 0.20508;\};\\
			$\begin{aligned}
			\qquad \texttt{consts} : \quad  &\textsl{A}= \texttt{null}, \\
			&\textsl{a} = \texttt{null}; \}\\
			\end{aligned}$\\
			entity \textsl{pump} : \textsl{Pump} \{ \\
			\qquad vars: \textsl{v} \{role: \textsl{exogenous};\};\\
			\qquad consts: \textsl{k} = null;\}\\
			\\
			process \textsc{inflow}(pump,tank1): \textsc{Inflow} \{\} \\
			process \textsc{valveTransmision}(tank1, tank2):\\
			\qquad\textsc{ValveTransmision} \{
			consts: \textsl{G}=4.429;\}  \\
			process \textsc{outflow}(tank2): \textsc{Outflow} \{
			consts: \textsl{G}=4.429;\}\\
			
			\hline
		\end{tabular}%
		\label{tab:IncompleteSS}%
	\end{table}%
	\vspace*{-\baselineskip}
\begin{table}[h]
	\caption{Modeling scenario with \twoPBM{}, for both the first stage (top) and the second stage (bottom). The initial values of $h_{1}$ and $h_{2}$ are taken from the measured data, while the parameter values of the upper tank in Stage 2 are the outcome of Stage 1. }
	\centering
	\ttfamily 
	\begin{tabular}{>{\footnotesize  }l}
		\hline
		//STAGE 1\\
		entity \textsl{tank1} : \textsl{Tank} \{ \\
		\qquad vars: \textsl{h} \{role: \textsl{endogenous}; initial: 0.38086;\};\\
		$\begin{aligned}
		\qquad \texttt{consts} : \quad  &\textsl{A}= \texttt{null}, \\
		&\textsl{a} = \texttt{null}; \}\\
		\end{aligned}$\\
		entity \textsl{pump} : \textsl{Pump} \{ \\
		\qquad vars: \textsl{v} \{role: \textsl{exogenous};\};\\
		\qquad consts: \textsl{k} = null;\}\\
		\\
		process \textsc{inflow}(pump,tank1): \textsc{Inflow} \{\} \\
		process \textsc{valveTransmision}(tank1):\\
		\qquad\textsc{ValveTransmision} \{
		consts: \textsl{G}=4.429;\}  \\
		\hline
	\end{tabular}%

\begin{tabular}{>{\footnotesize  }l}
	\hline
	//STAGE 2\\
	entity \textsl{tank1} : \textsl{Tank} \{ \\
	\qquad vars: \textsl{h} \{role: \textsl{exogenous}; initial: 0.38086;\};\\
	$\begin{aligned}
	\qquad \texttt{consts} : \quad  &\textsl{A}= 23.98, \\
	&\textsl{a} =0.762; \}\\
	\end{aligned}$\\
	entity \textsl{tank2} : \textsl{Tank} \{ \\
	\qquad vars: \textsl{h} \{role: \textsl{endogenous}; initial: 0.20508;\};\\
	$\begin{aligned}
	\qquad \texttt{consts} : \quad  &\textsl{A}= \texttt{null}, \\
	&\textsl{a} = \texttt{null}; \}\\
	\end{aligned}$\\

	process \textsc{valveTransmision}(tank1, tank2):\\
	\qquad\textsc{ValveTransmision} \{
	consts: \textsl{G}=4.429;\}  \\
	process \textsc{outflow}(tank2): \textsc{Outflow} \{
	consts: \textsl{G}=4.429;\}\\
	\hline
\end{tabular}%
	\label{tab:IncompleteMS}%
\end{table}%

\ifCLASSOPTIONcaptionsoff
  \newpage
\fi

\end{document}